\pdfoutput=1

\documentclass[11pt]{article}

\usepackage[]{acl}

\usepackage{times}
\usepackage{latexsym}

\usepackage[T1]{fontenc}

\usepackage[utf8]{inputenc}

\usepackage{microtype}

\usepackage{inconsolata}

\usepackage{graphicx}
\usepackage{amsmath}
\usepackage{multirow}
\usepackage{tabularx}
\usepackage{booktabs}
\usepackage[most]{tcolorbox}
\usepackage{xcolor}
\definecolor{PineGreen}{rgb}{0.0,0.47,0.44}
\definecolor{BrickRed}{rgb}{0.8, 0.25, 0.33}
\definecolor{ForestGreen}{rgb}{0.0, 0.57, 0.13}
\definecolor{Red}{rgb}{1.0, 0.0, 0.0}

\usepackage[subtle, tracking=normal, leading=normal]{savetrees}
\title{Multilingual Medical Reasoning \\ for Question Answering with Large Language Models}

\author{\\
 \textbf{Pietro Ferrazzi\textsuperscript{1,2}},
 \textbf{Aitor Soroa\textsuperscript{3}},
 \textbf{Rodrigo Agerri\textsuperscript{3}}
 \textsuperscript{1}Fondazione Bruno Kessler, Trento, Italy\\
 \textsuperscript{2}University of Padova, Italy \\
 \textsuperscript{3}HiTZ Center - Ixa, University of the Basque Country EHU\\
  \small{
    pferrazzi@fbk.eu, \{a.soroa,rodrigo.agerri\}@ehu.eus 
  }
}

\begin{document}
\maketitle
\begin{abstract}
Large Language Models with reasoning capabilities have recently shown strong promise for medical Question Answering. Current approaches report competitive results on English benchmarks through knowledge distillation from large models, yet multilingual coverage remains limited. We extend this line of research by presenting a method to generate multilingual reasoning traces that enrich the distillation paradigm with factual information from Wikipedia.
Using a retrieval-augmented generation pipeline over WikiProject Medicine, we produce 500k reasoning traces in English, Italian, and Spanish. The traces are generated to solve medical questions drawn from MedQA and MedMCQA, both of which we extend to Italian and Spanish. We evaluate our approach across in-domain and out-of-domain Medical QA benchmarks, demonstrating consistent performance gains in both few-shot in-context learning and supervised fine-tuning settings, achieving state-of-the-art results among 8B-parameter models across multiple model families. With the aim of fostering the development of multilingual clinical decision-support tools, we release the complete suite of resources: reasoning traces, translated datasets, Medical Wikipedia, and fine-tuned models.

\end{abstract}

\begin{figure*}[t]
  \includegraphics[width=\textwidth]{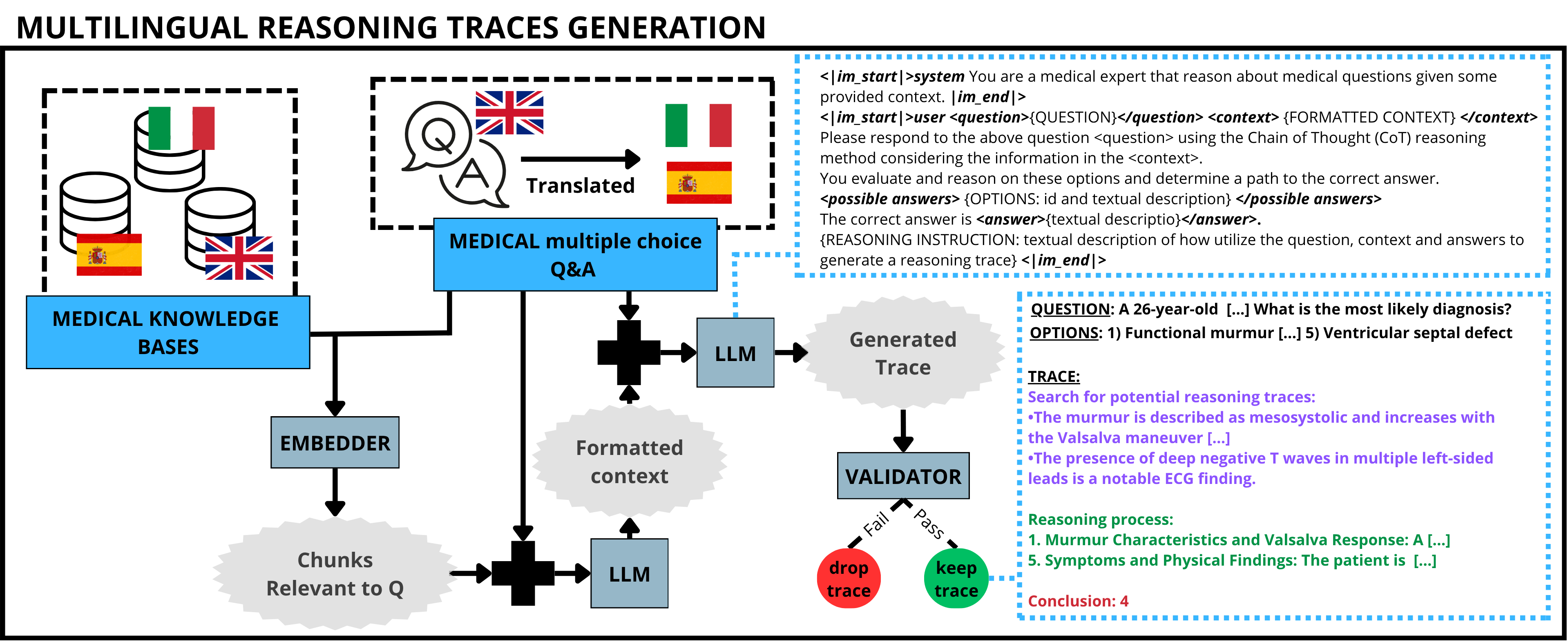}
  \caption{\textbf{Schema of the pipeline} to generate reasoning traces for multilingual multiple-choice medical question answering (QA). First, we create a Knowledge Base (KB) of medical information for each language. We extract relevant chunks from the KB for each pair of Question-Options (QO pair) in the source QA datasets, which we automatically port from English into Italian and Spanish. We prompt an LLM with the retrieved chunks and the QO pair for context formatting. Finally, we utilize the formatted context, the question, the options and the correct answer to generate a reasoning trace that answers the question itself. Answers that lead to the wrong conclusion are dropped, while the remaining form our reasoning traces dataset.
  The trace is composed of 3 parts. First (\textcolor{violet}{violet}), there is a search for potential paths and useful knowledge required to answer the question, which are then used to elaborate on the provided options (\textcolor{PineGreen}{green} part). Finally (\textcolor{BrickRed}{red}), the final answer is reported as conclusion.}
  \label{fig:pipeline}
\end{figure*}

\section{Introduction} \label{sec:intro}

Medical reasoning with Large Language Models (LLMs) is a rapidly evolving area aimed at improving Natural Language Processing systems in clinical settings.
Generating explicit reasoning traces, or chains of thought, has emerged as a promising approach to guide models toward more accurate 
answers \citep{NEURIPS2022_9d560961}. 
Moreover, the development of models explicitly trained to generate answers via complex, multi-step reasoning traces has marked a notable achievement for many domains \citep{plaat2025multistepreasoninglargelanguage}, while its direct application to
medicine presents unique challenges: 
the context is usually incomplete, 
concepts can be ambiguous, 
and there is a high degree of specialization. Such a scenario has fostered increased interest in medical reasoning, although a few important aspects remain underexplored. Recent approaches \citep{chen-etal-2025-towards-medical, huang2025m1unleashpotentialtesttime,yu2025finemedlmo} have primarily relied on distilling reasoning traces from powerful models to train smaller systems, suffering from 
heavy dependence on synthetic data.
Furthermore, multilingual reasoning has been largely neglected. Only \citet{chen-etal-2025-towards-medical} proposes a bilingual approach, including English and Chinese, that relies solely on distillation from OpenAI's models as a source of medical knowledge. While achieving good performance, this approach falls short in linking its reasoning to any sort of external medical knowledge.

In this paper, we propose a methodology to generate multilingual reasoning traces for multiple-choice medical question answering, based on the knowledge from the WikiProject Medicine\footnote{\url{https://en.wikipedia.org/wiki/Wikipedia:WikiProject_Medicine}}. 
We define reasoning traces as answer-conditioned rationales that elaborate medical notions and knowledge in order to reach a final answer to a question.
We evaluate such traces in various settings and demonstrate that training on them systematically yields state-of-the-art performance. Our contributions can be summarised as follows:
\begin{itemize}
    \item We present the first dataset of medical reasoning traces for Italian, Spanish, and English
    with the potential to be extended to any language in Wikipedia.
    \item We provide comprehensive experiments that demonstrate the effectiveness of applying our reasoning traces to multilingual, multiple-choice medical QA via in-context learning and supervised fine-tuning.
    \item We release a multilingual reasoning model that achieves state-of-the-art results in Italian, English, and Spanish medical QA.
\end{itemize}

In addition, we release two other datasets. First, a collection of all Wikipedia pages related to medicine in English, Italian, and Spanish. Second, a translated version of two common, English-based medical question answering datasets (MedMCQA and MedQA) in Italian and Spanish
that we use to guide and test the reasoning traces generation pipeline. All resources are available through HuggingFace\footnote{\url{huggingface.co/collections/NLP-FBK/multilngual-medical-reasoning}} and GitHub\footnote{\url{github.com/ferrazzipietro/multilingual-med-reasoner}}.

\section{Related work}\label{sec:related_work}


\noindent \textbf{Reasoning with LLMs.} In the context of Large Language Models, \textit{reasoning} refers to the ability to answer questions by complex, multi-step processes with intermediate steps \citep{li202512surveyreasoning}, producing a chain of thought before providing the actual answer to the user.
The first model explicitly designed for this purpose was OpenAI's GPT-o1 \citep{openai2024openaio1card}, 
whose development process remains mostly undisclosed.
Extensive work has been done on replicating GPT-o1 training phases and performance \citep{qin2024o1replicationjourneystrategic, huang2024o1replicationjourney, zeng2024scalingsearchlearningroadmap}. 
As a result of such efforts, DeepSeek's R1  \citep{deepseekai2025deepseekr1incentivizingreasoningcapability} was among the first open-source models to achieve comparable results, similarly to QwQ \citep{qwq32b}, marco-o1 \citep{zhao2024marcoo1openreasoningmodels}, and skywork-o1 \citep{he_2024_16998085}.
By achieving high performance in several benchmarks, these models have opened a line of research focused on the impact of reasoning in several domains, including math \citep{ahn-etal-2024-large}, physics \citep{xu2025ugphysics}, and biology \citep{bioprotocolbench2025}. 

\noindent \textbf{Reasoning for the medical domain.} Previous work showed how medical-oriented tuning of foundational LLMs can lead to major improvements \citep{luo2022biogpt, Wu2024}, even surpassing human experts on some benchmarks \citep{Singhal2025}. These results highlighted that domain-specific alignment can be beneficial for medical tasks. More recently, these findings have led to the integration of domain-shift and reasoning capabilities. 
Huatuo \citep{chen-etal-2025-towards-medical} was among the first released models, trained via distillation on medical questions. This approach was improved via test-time scaling by \citet{huang2025m1unleashpotentialtesttime}. \citet{wu2025medreasonelicitingfactualmedical} introduced the grounding of the distillation process in factual medical knowledge (more details in Appendix~\ref{ap:rel_work}).

\section{Methodology} \label{sec:methodology}

We introduce a novel methodology for distilling reasoning traces that are explicitly grounded in Wikipedia medical knowledge. This involves four steps: \textit{i)} selecting medical questions, \textit{ii)} constructing a curated Knowledge Base (KB) of reliable medical text, \textit{iii)} enriching each medical question with evidence retrieved from the KB using retrieval-augmented generation \citep{NEURIPS2020_6b493230}, and \textit{iv)} generating a step-by-step reasoning trace that leads to the correct answer among the given options with a high-capacity LLM. 

\subsection{Selection of Medical Questions}


\noindent \textbf{Datasets.} 
The MultiMedQA benchmark introduced in the Med-PaLM paper \citep{Singhal2023}, including MedMCQA \citep{pmlr-v174-pal22a}, MedQA \citep{medqa}, PubMedQA \citep{jin-etal-2019-pubmedqa}, and MMLU clinical \citep{hendrycks2021measuringmassivemultitasklanguage} is the main attempt to standardize multiple-choice medical QA datasets. 
Following prior work \citep{wu2025medreasonelicitingfactualmedical, chen-etal-2025-towards-medical}, we focus our approach on MedMCQA and MedQA, which are sets of exams from India and the USA.
Additionally, we include MedExpQA \citep{alonso2024medexpqa}, as it is currently the only manually validated multiple-choice medical QA dataset available for Italian, Spanish, and English. 
A summary of the selected datasets is provided in Appendix \ref{app:prompts_translation} (Table \ref{tab:dataset_descr}).

\noindent \textbf{Dataset translation.} MedMCQA and MedQA are English datasets. Given the multilingual objective of our work, we translate each Question-Options pair into Italian and Spanish using \texttt{Qwen-2.5-72B} \citep{qwen2025qwen25technicalreport}, prompted with a 5-shot example setup.
To assess the quality of the translation, we employ back-translation: the obtained text is translated back into English and compared against the original. 
We compared the original and back-translated question-answer pairs by means of COMET, chrF \citep{popovic-2015-chrf}, chrF++ \citep{popovic-2017-chrf}, and BERTScore \citep{zhang2019bertscore} metrics. 
Prior work has demonstrated that back-translation scores can serve as a useful proxy for translation quality and correlate with human judgments \citep{rapp-2009-backtranslation, zhuo-etal-2023-rethinking}.
Furthermore, the widespread use of back-translation as a data augmentation and validation technique \citep{sennrich-etal-2016-improving, bourgeade-etal-2024-data, sugiyama-yoshinaga-2019-data} supports its use for assessing translation quality in Italian and Spanish. 
As reported in Appendix~\ref{app:back_transl}, back-translation quality is consistently high across languages, achieving BERTScore of $96.6$, COMET $88.9$, chrF $78.6$, and chrF++ $77.9$.

As a further validation of the back-translation approach, we verify automatic metrics behavior on medical questions which translation has been manually revised (MedExpQA). We employ the method described above to translate English MedExpQA into Italian and Spanish. We compare the automatic translation to the manually revised translation using the same metrics employed for back-translation. The results (BERTscore $97$, COMET $88$, chrF $78$, chrF++ $75$) are consistent with what observed on MedQA and MedMCQA, supporting the hypothesis that these metrics indicate good alignment.

Since automatic metrics may not fully capture translation inaccuracies, we reserve the manually validated multilingual MedExpQA dataset exclusively for testing, guaranteeing out-of-domain evaluation on human-verified data.

\subsection{Knowledge Base Generation}

Our goal is to generate reasoning traces grounded in factual medical knowledge across the three languages. Hence, we require sources that comprehensively cover the range of medical specialties. 
Moreover, to ensure a fair comparison of results across languages, the underlying knowledge must be aligned across all three of them.
Accordingly, the requirements for the KB are: \textit{i)} reliability and diversity of the medical texts, and \textit{ii)} parallel information availability across the three languages. To satisfy these conditions, and taking advantage of its open collaborative editing model and of \emph{WikiProject Medicine}, we considered Wikipedia as our primary source of knowledge, based on \citet{smith2020situating}.

\noindent \textbf{Medical-Wikipedia creation.} \textcolor{black}{Although previous work on Medical QA has already tested Wikipedia as a medical knowledge source \citep{alonso2024medexpqa,xiong-etal-2024-benchmarking}, our method to leverage Wikipedia is novel.} We construct our medical knowledge base upon the WikiProject Medicine, a Wikipedia initiative that aims to collect all pages related to the medical domain.
First, we collected links to all the English-relevant pages\footnote{Pages listed at \url{en.wikipedia.org/wiki/Wikipedia:WikiProject_Medicine/Lists_of_pages/Articles}}.
For each page, we extracted the main text, infobox content, and interlanguage links. Using the links, we retrieved the corresponding pages in Italian and Spanish. 
This procedure results in a multilingual Medical-Wikipedia dataset in English, Italian, and Spanish, suitable for general NLP applications beyond medical question answering.

\noindent \textbf{From Medical-Wikipedia to Knowledge Base.} We employ the constructed dataset to build language-specific knowledge bases suitable for querying to retrieve context to support answering medical questions. To ensure a meaningful cross-lingual comparison, we enforce consistency across the three knowledge bases by retaining only pages available in all three languages. While this filtering step reduced the dataset size (particularly for English) it resulted in nearly parallel multilingual corpora. 
Although we cannot guarantee perfect alignment in page content and information density across languages, the resulting knowledge bases ensure that the same medical topics are present in all three languages.
We segmented the collected pages into chunks to construct the target knowledge bases. Each section of a page is treated as a chunk.
Sections exceeding 5,000 words were discarded, while sections containing fewer than 250 words were merged with the preceding ones. We filtered out irrelevant sections, such as ``Bibliography'' and ``External links'' (a comprehensive list is provided in Appendix \ref{app:wikipedia_excluded_sections}). An overview of the outcome of each step is presented in Table~\ref{tab:medical_wiki}.

\begin{table}
  \centering
  \small
\setlength{\tabcolsep}{4.5pt}
  \renewcommand{\arraystretch}{1.2}
  \begin{tabular}{l|rrrr}
    \toprule
     & \multicolumn{3}{c}{\textbf{Number of pages (million words)}} \\
      & \multicolumn{1}{c}{\textbf{Source}} & \multicolumn{1}{c}{\textbf{De-duplicated}} & \multicolumn{1}{c}{\textbf{All 3 lang}} \\
    \midrule
    en & 33.859 (81) & 29.306 (72) & 8.577 (20) \\
    it & 11.045 (15) & 8.492  (8) & 8.492 (8) \\
    es & 12.886 (20) & 10.124 (15) & 8.513 (11)  \\
    \bottomrule
  \end{tabular}
  \caption{\label{pages-stats}
    \textbf{Pages extracted from the Wikipedia Project} for the three languages (\textit{Source}). \textit{De-duplication} combines medical topics repeated in multiple links. 
    Finally, the KB is restricted to pages available in all three languages, with slight variations in page counts, as concepts can be partitioned differently across pages in different languages.
  }
  \label{tab:medical_wiki}
\end{table}

\subsection{Retrieval Augmentation}

Once the knowledge base is constructed, we enriched each medical question from MedQA and MedMCQA with the most relevant contextual information. Following the paradigm of Retrieval-Augmented Generation, we pre-pended the retrieved passages to the model input prior to prompting (Figure~\ref{fig:pipeline}). We pre-compute the embedding of each chunk in the three language-specific knowledge bases, calculating their cosine similarity with each Question-Options pair. We select the top five most similar chunks as context. Embeddings are computed using the top-performing model from the MTEB leaderboard \citep{enevoldsen2025mmtebmassivemultilingualtext} retrieval task at the time of writing (\texttt{Qwen3-Embedding-8B} by \citet{zhang2025qwen3embeddingadvancingtext}). 
To address the issue of potential redundancy and inclusion of irrelevant information in the context highlighted by \citet{wang-etal-2024-searching}, we prompt an LLM (\texttt{Qwen3-32B}, \citet{yang2025qwen3technicalreport}) to rewrite the retrieved chunks. This step ensures consistency within the retrieved chunks, enhancing conciseness and avoiding duplication. The prompt template and more details on the results in Appendix~\ref{app:prompts}.

\subsection{Generation of Reasoning Traces}

We prompt an open-source reasoning model, \texttt{Qwen3-32B}, to generate reasoning traces, providing an input composed of four parts
: \textit{i)} the formatted contextual information retrieved from the KB, to provide factual knowledge on the medical topic; \textit{ii)} the medical question that defines the problem \textit{iii) }the answer options that constrain the space of reasoning paths; \textit{iv)} the correct answer to guide the model toward selecting the most appropriate path. 
We enforce the use of the correct answer, motivated by our goal of generating the most accurate and informative traces possible. To prevent the model from collapsing into simply stating the correct answer, we design the prompt to encourage exploration of a space of potential answers before reaching the conclusion. 
The system prompt, with a structure based on the findings of \citet{wu2025medreasonelicitingfactualmedical}, is described in Appendix~\ref{app:prompts}.
We verify the traces' conclusions by extracting the answer identifier via regular expression matching and discarding the incorrect ones, as suggested by the ``philosophy of verifiability'' in \citet{chen-etal-2025-towards-medical}. 
Notably, this resulted in discarding less than 5\% of the traces, meaning that our pipeline produces answers that usually lead to correct answers.

\subsection{Multilingual Reasoning Traces Dataset}

We obtain a dataset of more than $500k$ reasoning traces, each answering a different medical question given the options in one of the three languages.
The Italian split consists of $166.257$ traces from MedMCQA, $9.468$ from MedQA; the Spanish split of $168.771$ from MedMCQA, $9.584$ from MedQA;
the English split of $169.098$ from MedMCQA, $9.520$ from MedQA.
An example of a generated trace can be seen in the Appendix, Figure~\ref{fig:example}.

\begin{table*}[!h]
  \centering
  \setlength{\tabcolsep}{4.5pt}
  \renewcommand{\arraystretch}{1.2}
  \scriptsize
  \begin{tabular}{l|ccc|ccc|ccc|c}
    \toprule
     & \multicolumn{3}{c|}{\textbf{MedExpQA}} 
     & \multicolumn{3}{c|}{\textbf{MedMCQA}} 
     & \multicolumn{3}{c|}{\textbf{MedQA}} & \textbf{AVG}\\
    \hline
    \textbf{model} & \textbf{IT} & \textbf{EN} & \textbf{ES}& \textbf{IT} & \textbf{EN} & \textbf{ES} & \textbf{IT} & \textbf{EN} & \textbf{ES} &\\
    \midrule
    Llama3-1B &
    26.3  \textcolor{ForestGreen}{+2.6} & 
    33.0  \textcolor{BrickRed}{-1.2} & 
    27.3  \textcolor{ForestGreen}{+0.0} & 
    31.4  \textcolor{ForestGreen}{+0.6} & 
    40.2  \textcolor{ForestGreen}{+3.8} & 
    33.6  \textcolor{ForestGreen}{+2.6} & 
    27.4  \textcolor{ForestGreen}{+0.1} & 
    35.9  \textcolor{ForestGreen}{+2.5} & 
    32.0  \textcolor{ForestGreen}{+3.5} &
    31.9  \textcolor{ForestGreen}{+1.6}\\
    \midrule
    Qwen3-1.7B &
    45.1  \textcolor{ForestGreen}{+8.1} & 
    51.9  \textcolor{ForestGreen}{+3.6} & 
    41.6  \textcolor{ForestGreen}{+4.7} & 
    51.5  \textcolor{ForestGreen}{+16.9} & 
    57.2  \textcolor{ForestGreen}{+19.8} & 
    53.4  \textcolor{ForestGreen}{+18.0} & 
    46.8  \textcolor{ForestGreen}{+11.3} & 
    57.2  \textcolor{ForestGreen}{+16.7} & 
    48.3  \textcolor{ForestGreen}{+13.4}  &
    50.3  \textcolor{ForestGreen}{+12.5}\\
    \midrule
    Gemma3-4B &
    47.8  \textcolor{BrickRed}{-1.9} & 
    52.5  \textcolor{ForestGreen}{+11.8} & 
    52.5  \textcolor{BrickRed}{-3.1} & 
    54.1  \textcolor{ForestGreen}{+11.4} & 
    58.7  \textcolor{ForestGreen}{+11.9} & 
    55.1  \textcolor{ForestGreen}{+12.3} & 
    50.7  \textcolor{ForestGreen}{+7.2} & 
    62.9  \textcolor{ForestGreen}{+13.4} & 
    53.5  \textcolor{ForestGreen}{+9.1}  &
    54.2  \textcolor{ForestGreen}{+8.0}\\
    \midrule
    MedGemma-4B &
    53.8  \textcolor{BrickRed}{-2.7} & 
    61.8  \textcolor{ForestGreen}{+3.1} & 
    59.2  \textcolor{ForestGreen}{+5.6} & 
    56.5  \textcolor{ForestGreen}{+9.9} & 
    62.5  \textcolor{ForestGreen}{+10.7} & 
    57.9  \textcolor{ForestGreen}{+11.1} & 
    55.1  \textcolor{ForestGreen}{+4.6} & 
    67.2  \textcolor{ForestGreen}{+11.8} & 
    57.0  \textcolor{ForestGreen}{+7.1}   &
    59.0  \textcolor{ForestGreen}{+6.8}\\
    \midrule
    Llama3-8B &
    52.4  \textcolor{ForestGreen}{+3.2} & 
    69.4  \textcolor{ForestGreen}{+6.4} & 
    50.7  \textcolor{ForestGreen}{+2.8} & 
    55.7  \textcolor{ForestGreen}{+9.4} & 
    63.4  \textcolor{ForestGreen}{+6.0} & 
    58.0  \textcolor{ForestGreen}{+10.1} & 
    53.4  \textcolor{ForestGreen}{+3.0} & 
    70.9  \textcolor{ForestGreen}{+7.8} & 
    57.4  \textcolor{ForestGreen}{+7.6}   &
    59.0  \textcolor{ForestGreen}{+6.2}\\
    \midrule
    Qwen3-8B &
    64.7  \textcolor{ForestGreen}{+4.6} & 
    74.6  \textcolor{ForestGreen}{+8.6} & 
    64.8  \textcolor{ForestGreen}{+0.0} & 
    62.4  \textcolor{ForestGreen}{+9.3} & 
    67.4  \textcolor{ForestGreen}{+8.8} & 
    64.3  \textcolor{ForestGreen}{+10.4} & 
    67.5  \textcolor{ForestGreen}{+10.5} & 
    75.6  \textcolor{ForestGreen}{+8.5} & 
    66.5  \textcolor{ForestGreen}{+6.4}   &
    67.5  \textcolor{ForestGreen}{+7.5}\\
   \midrule
    Gemma3-27B &
    74.9  \textcolor{BrickRed}{-7.0} & 
    79.4  \textcolor{BrickRed}{-0.3} & 
    77.7  \textcolor{BrickRed}{-4.5} & 
    67.3  \textcolor{ForestGreen}{+9.4} & 
    71.0  \textcolor{ForestGreen}{+8.2} & 
    68.4  \textcolor{ForestGreen}{+8.9} & 
    75.2  \textcolor{ForestGreen}{+6.0} & 
    79.8  \textcolor{ForestGreen}{+6.9} & 
    74.4  \textcolor{ForestGreen}{+5.5}  &
    74.2  \textcolor{ForestGreen}{+3.7}\\
   \midrule
    MedGemma-27B &
    76.8  \textcolor{BrickRed}{-3.4} & 
    84.6  \textcolor{ForestGreen}{+4.5} & 
    81.9  \textcolor{BrickRed}{-2.7} & 
    70.4  \textcolor{ForestGreen}{+8.7} & 
    76.0  \textcolor{ForestGreen}{+7.7} & 
    72.5  \textcolor{ForestGreen}{+9.8} & 
    82.4  \textcolor{ForestGreen}{+9.6} & 
    87.1  \textcolor{ForestGreen}{+9.5} & 
    81.5  \textcolor{ForestGreen}{+8.8} &
    79.2  \textcolor{ForestGreen}{+5.8}\\
    \midrule
    Qwen3-32B &
    77.6  \textcolor{ForestGreen}{+0.2} & 
    80.6  \textcolor{BrickRed}{-2.1} & 
    81.9  \textcolor{ForestGreen}{+0.8} & 
    69.6  \textcolor{ForestGreen}{+5.7} & 
    74.7  \textcolor{ForestGreen}{+5.8} & 
    70.9  \textcolor{ForestGreen}{+5.7} & 
    76.4  \textcolor{ForestGreen}{+4.1} & 
    84.3  \textcolor{ForestGreen}{+7.8} & 
    78.4  \textcolor{ForestGreen}{+4.5} &
    77.1  \textcolor{ForestGreen}{+3.6}\\
    \midrule
    Llama3-70B &
    78.9  \textcolor{ForestGreen}{+4.9} & 
    83.8  \textcolor{ForestGreen}{+1.8} & 
    82.2  \textcolor{ForestGreen}{+5.2} & 
    71.1  \textcolor{ForestGreen}{+5.7} & 
    76.0  \textcolor{ForestGreen}{+4.2} & 
    73.3  \textcolor{ForestGreen}{+6.9} & 
    80.0  \textcolor{ForestGreen}{+14.7} & 
    85.5  \textcolor{ForestGreen}{+13.7} & 
    80.0  \textcolor{ForestGreen}{+13.7} &
    79.0  \textcolor{ForestGreen}{+7.9}\\
    \bottomrule
    \toprule
    AVG $\delta$ 
    &
    \textcolor{ForestGreen}{+0.9} &
    \textcolor{ForestGreen}{+3.6} &
    \textcolor{ForestGreen}{+0.9} &
    \textcolor{ForestGreen}{+8.7} &
    \textcolor{ForestGreen}{+8.7} &
    \textcolor{ForestGreen}{+9.6} &
    \textcolor{ForestGreen}{+7.1} & 
    \textcolor{ForestGreen}{+9.9} &
    \textcolor{ForestGreen}{+8.0} &
    \textcolor{ForestGreen}{+5.8} \\    
    \bottomrule
    \toprule
    GPT-4o  &
    87.0  \textcolor{BrickRed}{-2.6} & 
    88.1  \textcolor{ForestGreen}{+2.9} & 
    89.9  \textcolor{ForestGreen}{+2.4} & 
    87.2  \textcolor{ForestGreen}{+1.1} & 
    91.2  \textcolor{ForestGreen}{+1.8} & 
    88.5  \textcolor{ForestGreen}{+0.8} & 
    75.0  \textcolor{BrickRed}{-0.5} & 
    79.1  \textcolor{BrickRed}{-1.4} & 
    76.0  \textcolor{BrickRed}{-0.4} & 
    84.7  \textcolor{ForestGreen}{+0.5} \\
    \midrule
  \end{tabular}
  \caption{\label{tab:main_results_fs}
  \textbf{Performances of models exposed to our traces via in-context learning} (2-shot examples, each being a question-trace-answer triplet). Accuracy is reported with deltas ($\delta$) from baseline (2-shot examples without traces).
  }
\end{table*}

\begin{table*}[t]
  \centering
  \setlength{\tabcolsep}{4.5pt} 
  \renewcommand{\arraystretch}{1.2} 
  \scriptsize
  \begin{tabular}{p{1.5cm}|ccc|ccc|ccc|c}
    \toprule
     & \multicolumn{3}{c|}{\textbf{MedExpQA}} & \multicolumn{3}{c|}{\textbf{MedMCQA}} & \multicolumn{3}{c|}{\textbf{MedQA}} & \textbf{AVG $\delta$} \\
    \hline
    \textbf{model} & \textbf{IT} & \textbf{EN} & \textbf{ES} & \textbf{IT} & \textbf{EN} & \textbf{ES} & \textbf{IT} & \textbf{EN} & \textbf{ES} & \\
    \midrule
    Llama3-8B   & 
    58.0 \textcolor{ForestGreen}{+9.0} & 
    65.4 \textcolor{ForestGreen}{+2.7} & 
    54.3 \textcolor{ForestGreen}{+6.4} & 
    46.8 \textcolor{ForestGreen}{+0.4} & 
    59.0 \textcolor{ForestGreen}{+1.5} & 
    50.6 \textcolor{ForestGreen}{+2.7} & 
    54.0 \textcolor{ForestGreen}{+2.3} & 
    71.6 \textcolor{ForestGreen}{+4.4} & 
    59.1 \textcolor{ForestGreen}{+6.2} &
    57.6 \textcolor{ForestGreen}{+4.0}\\
    Qwen3-8B  & 
    62.2 \textcolor{ForestGreen}{+2.1} & 
    69.7 \textcolor{ForestGreen}{+3.7} & 
    69.7 \textcolor{ForestGreen}{+4.8} & 
    52.3 \textcolor{BrickRed}{-0.7} & 
    60.9 \textcolor{ForestGreen}{+2.3} & 
    53.2 \textcolor{BrickRed}{-0.6} & 
    61.7 \textcolor{ForestGreen}{+3.1} & 
    71.7 \textcolor{ForestGreen}{+5.4} & 
    65.0 \textcolor{ForestGreen}{+4.0} &
    63.0 \textcolor{ForestGreen}{+2.7}\\
     \toprule
    AVG $\delta$ 
    &
    \textcolor{ForestGreen}{+5.6} &
    \textcolor{ForestGreen}{+3.2} &
    \textcolor{ForestGreen}{+5.6} &
    \textcolor{BrickRed}{-0.1} &
    \textcolor{ForestGreen}{+1.9} &
    \textcolor{ForestGreen}{+1.0} &
    \textcolor{ForestGreen}{+2.7} & 
    \textcolor{ForestGreen}{+4.9} &
    \textcolor{ForestGreen}{+5.1} & 
    \textcolor{ForestGreen}{+3.3}\\
    \bottomrule
  \end{tabular}
  \caption{\label{tab:ft_our}
  \textbf{Performances of models fine-tuned on our reasoning traces} (SFT). Accuracy is reported with deltas from the baseline (2-shot without our reasoning traces).}
\end{table*}

\section{Traces Evaluation}\label{sec:results}
We evaluate the usefulness of our generated reasoning traces by measuring how effectively they support Medical QA in both in-context learning (ICL) and supervised fine-tuning (SFT) settings. Our primary metric is downstream task accuracy, which directly reflects whether the traces fulfill their intended purpose. Nonetheless, in Section \ref{sec:error-analysis} we offer a manual evaluation of our newly generated reasoning traces by having 11 medical doctors analyze the correctness of a randomly chosen sample of \textit{question, option, reasoning} triplets.


For both ICL (few-shot) and SFT experiments, examples are drawn from the train splits of MedQA and MedMCQA. Out-of-domain evaluation is conducted on the combined train, validation, and test splits of MedExpQA, guaranteeing out-of-distribution testing on human-validated multilingual data. An additional evaluation is also performed on the test splits of MedQA and MedMCQA (Appendix \ref{app:prompts_translation}, Table~\ref{tab:dataset_descr}), guaranteeing that the models do not, at any time in the evaluation step, see the correct answers from the training split used to generate the reasoning traces. Finally, we also report a comparison with previous state-of-the-art.

\noindent \textbf{Baselines.} To determine if our traces are helpful, we need to define a baseline to compare against. 
We adopt a few-shot evaluation setting, where each prompt includes two example questions with their options and correct answers, followed by the test question. To ensure a competitive baseline, the examples are selected through similarity search in an embedding space: for each test question, we retrieve the most similar training questions (along with their answers) to use as few-shot examples.

\noindent \textbf{Statistical analysis}
We employ statistical tests to determine the significance of the differences in performance. For each comparison, we treat the combination of model, dataset, and language as an independent unit, representing a single instance of the treatment (i.e., exposure to the reasoning traces). We test the hypothesis that the treatment produces an effect greater than zero. Since multiple statistical tests are performed on the same data, we apply a Bonferroni correction to adjust the significance threshold.

\subsection{In-Context Learning} \label{sec:eval_fs}

We determine the impact of utilizing our traces via ICL. While baseline examples include only questions, options, and answers, our ICL setup also includes the reasoning traces in the $2$ examples. The retrieval strategy remains identical across both settings, ensuring that any observed improvements can be solely attributed to the inclusion of our traces.
We test several models from diverse families in their instructed versions and report the results in Table~\ref{tab:main_results_fs}.

\begin{figure*}[t]
  \includegraphics[width=0.48\linewidth]{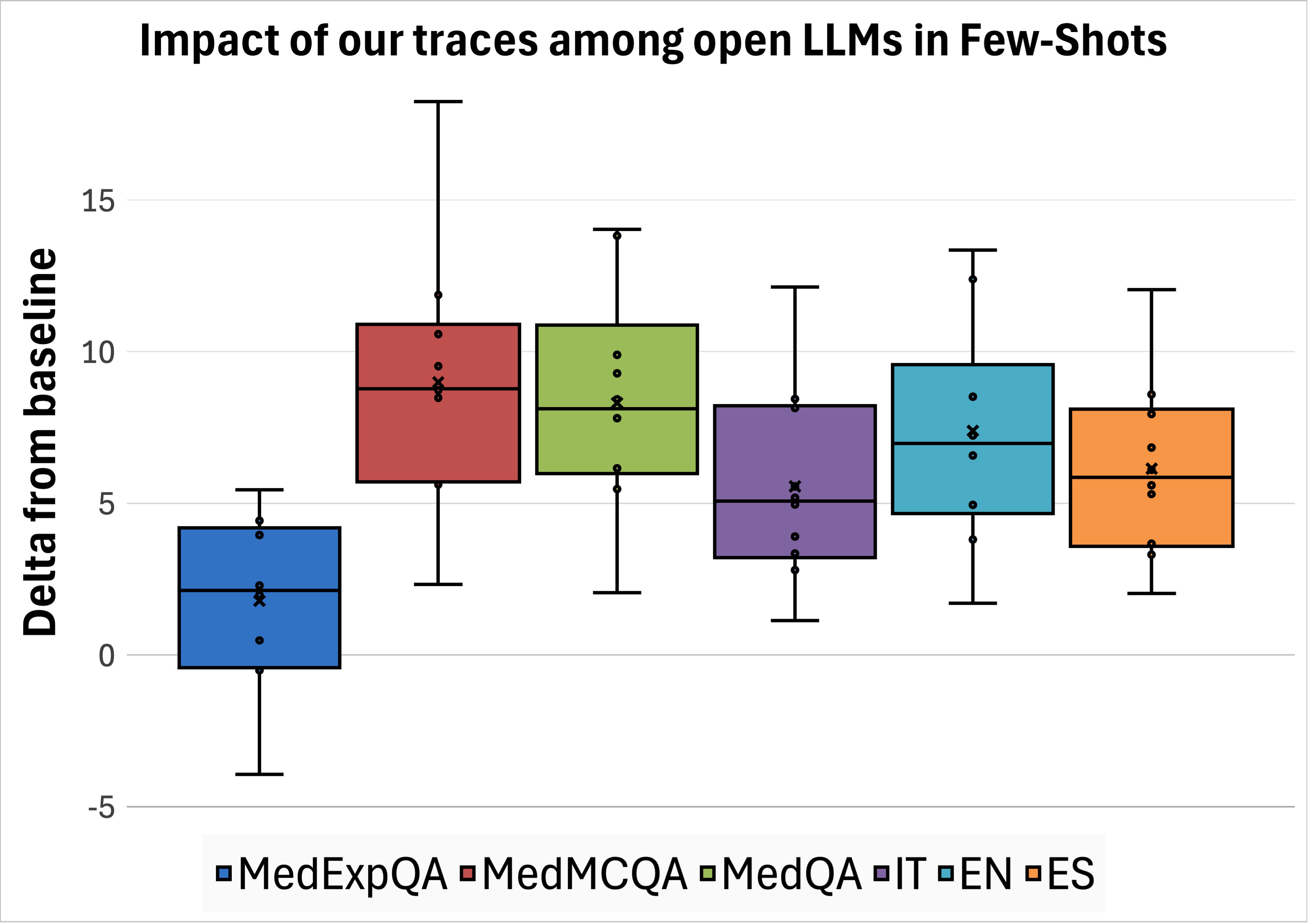} \hfill
  \includegraphics[width=0.48\linewidth]{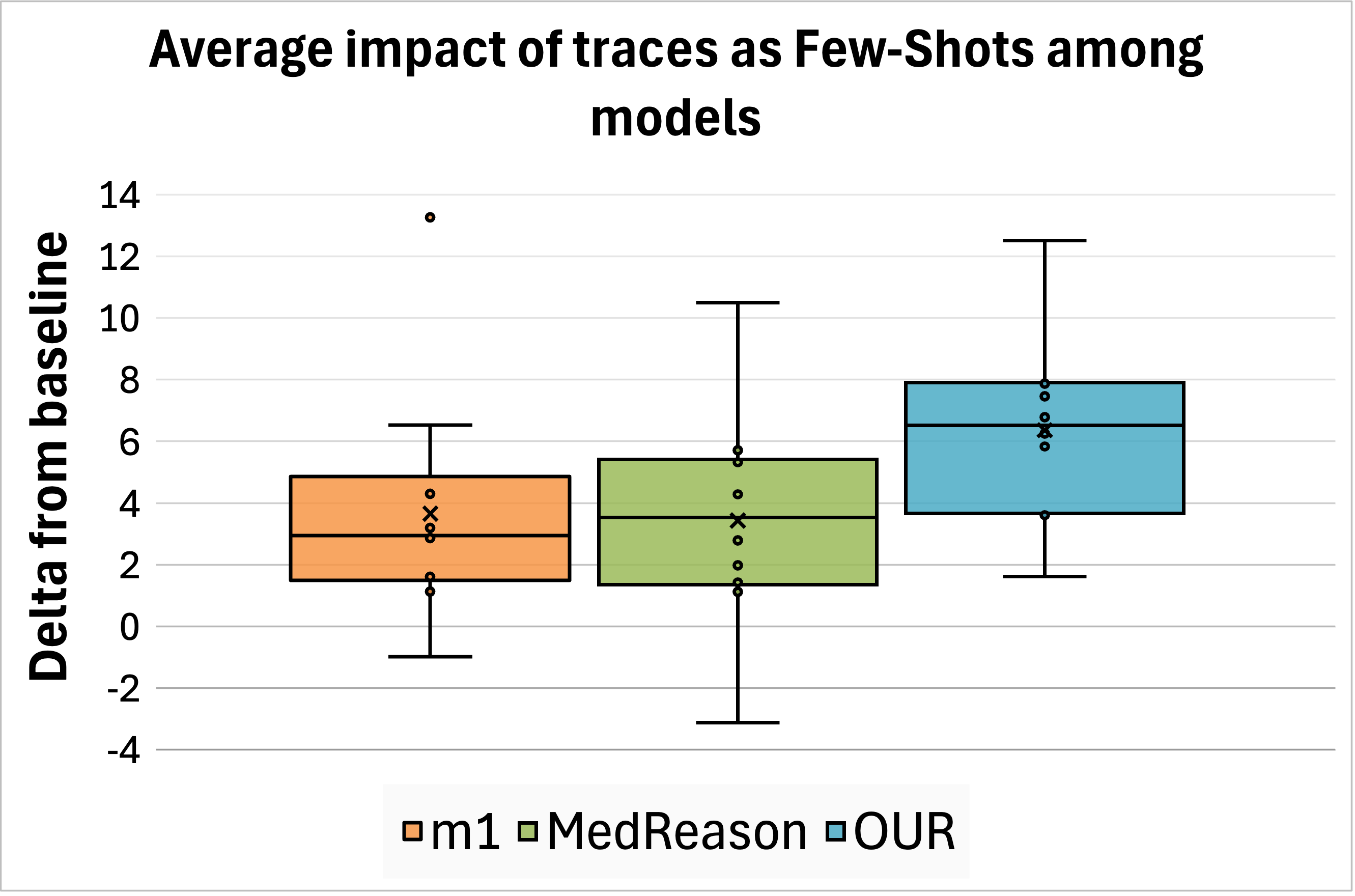} \hfill
  \includegraphics[width=0.48\linewidth]{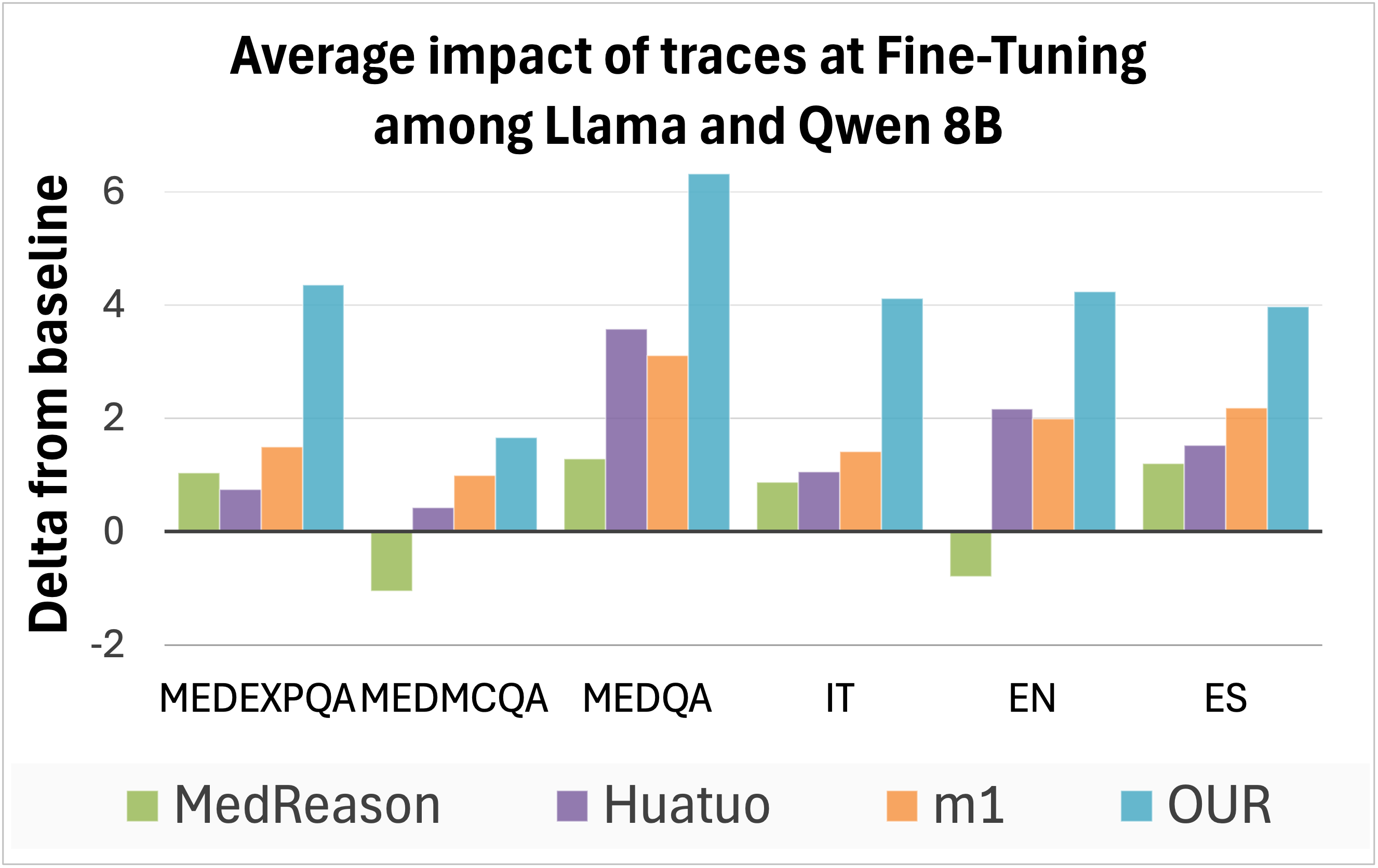} \hfill
  \includegraphics[width=0.48\linewidth]{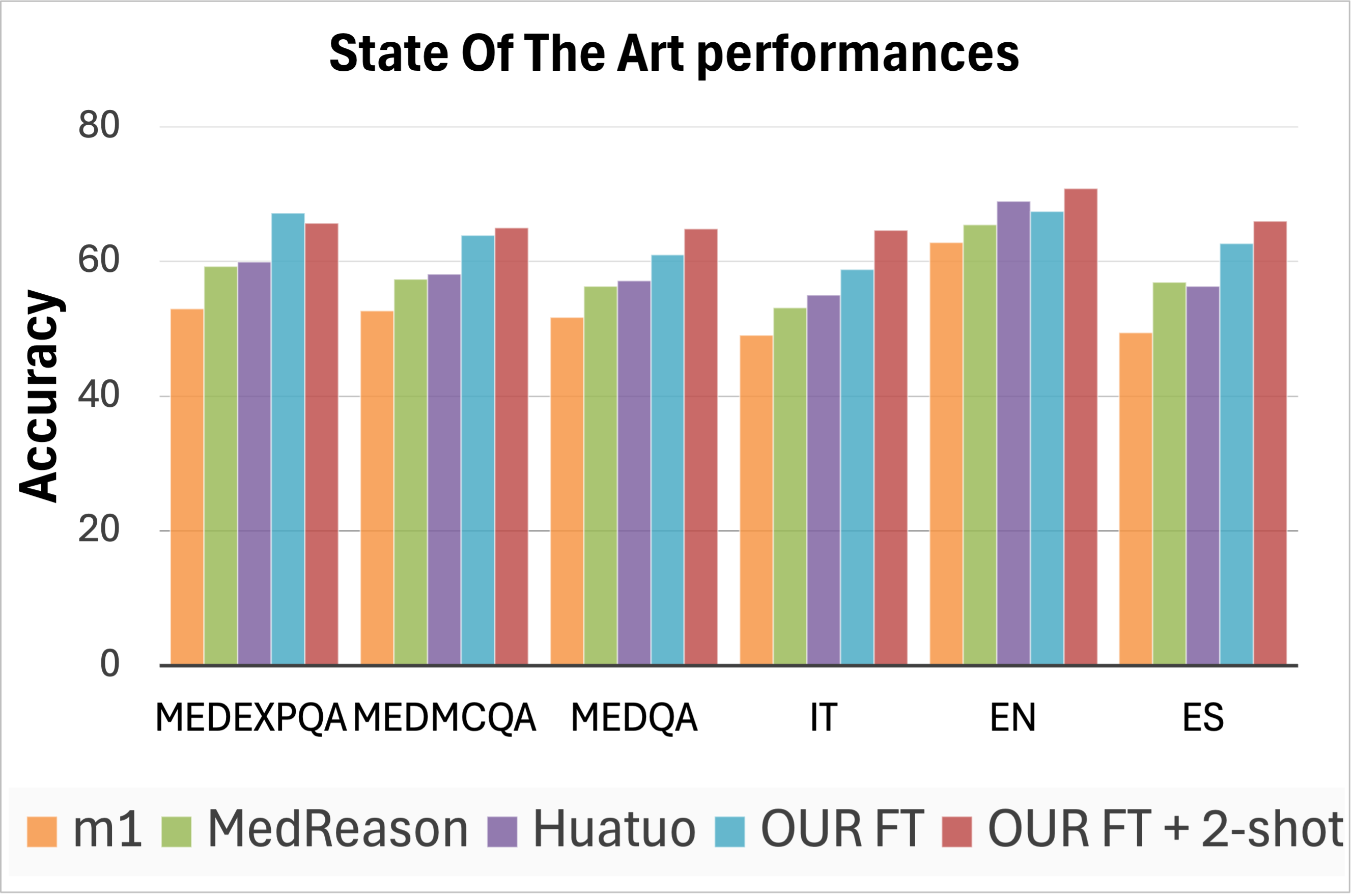} \hfill
  \caption {\textbf{Impact of our traces via  In-Context Learning (ICL) and Supervised Fine-Tuning (SFT)}. 
  The \textbf{top-left} boxplots represent the improvement due to exposing the $10$ open-source models to our traces as two-shot examples (\textit{ICL}). The improvement is measured in terms of increased accuracy over the baseline (model prompted with two-shot, where each example is a question-answer pair without the reasoning trace).
  The \textbf{top-right} box-plots represent the improvement in accuracy among $10$ open-source models by trace type when provided at inference time (two-shot) with respect to the baseline. Our traces are more effective than previously proposed ones. 
  The \textbf{bottom-left} histogram shows the average impact of different types of traces when provided at fine-tuning time. 
  The \textbf{bottom-right} histogram shows the performances of the best models proposed in the literature against our fine-tuned version of \texttt{Qwen3-8B} ("OUR FT") and the same model further enhanced with $2$ of our traces \textcolor{black}{("OUR FT + 2-shot")}.} 
  \label{fig:comp}
\end{figure*}

\noindent \textbf{Results.} We observe that our traces enable all models to generate more accurate answers compared to the baseline, as reported in Table~\ref{tab:main_results_fs}. The average effect of our traces among datasets and languages is shown in Figure~\ref{fig:comp} (top-right).
For medical questions in MedMCQA and MedQA, we observe an average increase in accuracy of $+7$ to $+10$ points across all languages. The out-of-distribution dataset (MedExpQA) is the one that benefits the least, with the case of Gemma-3-27B and Qwen-3-32B not getting any benefit. 
Nevertheless, the overall impact among models on this dataset is positive, with an average increase in accuracy of $+1.8$ points. 
To verify its statistical significance, we apply a paired t-test on the performance deltas between models trained with and without our traces, obtaining a p-value of 0.02, which falls below the standard significance threshold of $\alpha$ = 0.05.

\noindent \textbf{Disentangling reasoning from knowledge.}
When we expose models to our traces via ICL, we select the two examples that are the most similar to the target. These example traces therefore include medical knowledge that is required to answer the test question. By removing the similarity-based selection of the examples -- and picking them randomly instead -- we can estimate how much of the increase in performance is due to the reasoning capabilities acquired via ICL by itself, disentangling the effect of the similarity-based example traces.

For computational constraints, we re-run all ICL with 2 random examples on 9/10 models (excluding Llama-70B), MedQA (in-distribution), and the test set of MedExpQA (out-of-distribution)\footnote{While deltas for MedExpQA are comparable with what reported in Table~\ref{tab:fs_all}, the absolute numbers are not, as here we run inference solely on the test set.}. 
Randomly picked examples have a positive impact on 8/9 models (Llama-1B, the weakest model by a margin), with an average impact of +2.5 points (MedExpQA), and +4.7 (MedQA). This result highlights that reasoning capabilities acquired via ICL are useful for the models, even when the traces they are exposed to are not selected to be the most relevant for the test question.
Full results are reported in the Appendix (Table~\ref{tab:main_results_fs_no_rag}).

\subsection{Supervised Fine-Tuning} \label{sec:ft}

To measure how our reasoning traces influence model learning, we fine-tuned models in a supervised setting. Due to computational constraints, we restricted our experiments to \texttt{Llama3} and \texttt{Qwen3} in their \texttt{8B} versions. We trained each model on examples formatted as 
\textit{$<$q$>$\{question\}$<$/q$>$$<$o$>$\{options\}$<$/o$>$\{trace\}}.

We follow \citet{thapa2025disentangling} by using ``reasoning intensive'' examples for training as opposed to ``knowledge intensive'', and \citet{liu2025distillationpushinglimitsmedical}, which suggests prioritizing MedQA sampling ratio over MedMCQA. 
Our final fine-tuning dataset comprises 5,837 traces from MedMCQA and 5,594 from MedQA per language, for a total of 34,293 examples across English, Italian, and Spanish.

We fine-tuned the models for $3$ epochs on two H200 GPUs, with a per-device batch size of $32$, AdamW optimizer with a learning rate of $5e-6$, a cosine scheduler, and a warm-up ratio of $0.1$. Training on these settings took $1.5$ hours per model. 

\noindent \textbf{Results.} We observed that training on our traces benefits both \texttt{Llama-3.1-8B-Instruct} and \texttt{Qwen3-8B}, yielding an average accuracy improvement of $+3.9$ points when compared with the baseline (Table~\ref{tab:ft_our}). 
The average impact is significantly higher than zero with a p-value $< 0.005$.
Interestingly, for \textit{in-distribution} datasets (MedMCQA, MedQA), the gains are smaller than those seen in ICL settings. We attribute this to the nature of the traces selected as two-shot examples: they are highly relevant for the test question, acting as targeted guidance for instructed models that are capable of handling long prompts.
In contrast, fine-tuning aims to generalize such knowledge, which turns out to be less effective than delivering the exact relevant information at inference time. 
We further analyze this behavior in the next section, showing that combining fine-tuned models with our traces in few-shot prompts leads to the best overall performance.
On the other hand, the opposite is observed for \textit{out-of-distribution} performance (MedExpQA), where SFT obtained a higher average increase among languages with respect to the baseline (+4.8 against +1.8 in ICL). This is reasonable, as SFT enables the model to internalize reasoning patterns and medical knowledge that generalize beyond the specific examples provided at inference time. In contrast, ICL relies on retrieving highly relevant in-context examples, which may be less aligned with out-of-distribution questions.

\vspace{1cm}

\subsection{Comparison with previous work} \label{sec:comparison}


As described in Section~\ref{sec:related_work}, the main previous efforts to generate reasoning traces for medical question answering are Huatuo \citep{chen-etal-2025-towards-medical}, MedReason \citep{wu2025medreasonelicitingfactualmedical}, and m1-m23k \citep{huang2025m1unleashpotentialtesttime}. They released $19.704$, $32.682$, and $23.493$ question-trace pairs (licence \texttt{apache 2.0}), respectively, on which they trained models using SFT and reinforcement learning techniques.
We aim to compare our traces against those from previous work. First, we evaluate their effect in \textit{i)} few-shot prompting and \textit{ii)} supervised fine-tuning. Finally, \textit{iii)} we evaluate each method under its strongest reported configuration. This final step goes beyond a trace-level comparison, demonstrating that our best model consistently outperforms any other alternatives.

We do not perform a comparison to the generation mechanisms presented in the literature, but to the data resources existing work released \footnote{We employ the reasoning traces as in the official releases in HuggingFace: \texttt{UCSC-VLAA/MedReason},\texttt{ UCSC-VLAA/m23k-tokenized}, and \texttt{FreedomIntelligence/medical-o1-reasoning-SFT}.}. A summary of the characteristics of the traces we employ is reported in Table~\ref{tab:summary_traces_rel}. Our traces' length is comparable to MedReason ($226$ vs. $370$ words on average) and substantially shorter than m1 ($1123$ tokens), while providing a much larger pool of ICL examples ($179k$ vs. $32.7k$ for MedReason and $23.5$k for m1). While m1 and Huatuo use GPT-4o for the generation, m1 and us use open-source models (DeepSeek R1, Qwen3-32B). All traces are generated from MedQA and MedMCQA.
For SFT, we downsample to $34k$ examples, making our dataset comparable to previous monolingual approaches for English. As our traces are shorter, this corresponds to $7.8M$ words in English and $23M$ words across the three languages, comparable to m1 ($26M$ words) and larger than Huatuo ($16M$ words) and MedReason ($12M$). 

\begin{table}[]
\centering
\small
\begin{tabular}{l|p{0.9cm}p{0.9cm}|p{0.9cm}p{0.9cm}} \toprule
 & \multicolumn{2}{c|}{\textbf{ICL}} & \multicolumn{2}{c}{\textbf{SFT}} \\ \midrule
 & \textbf{n traces} & \textbf{avg lenght} & \textbf{n traces} & \textbf{tot words} \\ \midrule
MedReason & 32.7 k & 370 & 32.7 k & 12.1M \\
m1 & 23.5 k & 1123 & 23.5 k & 26.4M \\
Huatuo-o1 & - & - & 40k & 16.0M \\
OUR-en & 179k & 228 & 34k & 7.8M \\
OUR-en+es+it & 530k & 226 & 102k & 23.0M   \\
\bottomrule
\end{tabular}
\caption{\textbf{Summary of the characteristics of each trace type}. Comparison of the number of traces and average trace length (words) for In-Context Learning (ICL) and the total number of traces and words for Supervised Fine-Tuning (SFT) for the different trace types. MedReason, m1, and Huatuo are English-only. Therefore, we report statistics on our approach for both English and multilingual traces. Huatuo is not employed via ICL.}\label{tab:summary_traces_rel}
\end{table}

\noindent \textbf{Few-Shot.} We test the effect of \textit{MedReason} and \textit{m1} traces at inference time on the same eleven models described in Section~\ref{sec:eval_fs} and measure accuracy improvements, directly comparing against our traces. \textit{Huatuo} traces could not be evaluated in this setting because they are framed as open-ended questions without multiple-choice options or unique correct answers. We observed that our traces provide the greatest average improvement among datasets and languages; overall impact in Figure~\ref{fig:comp} (top-right) (more details in Appendix ~\ref {app:comp}).

\noindent \textbf{Fine-Tuning.} We fine-tune the \texttt{8B} versions of \texttt{Llama3} and \texttt{Qwen3} using the reasoning traces and official codebases provided by \textit{MedReason}, \textit{m1}, and \textit{Huatuo}. We retained the hyperparameters specified in each work, training on two H200 GPUs for a total of approximately 18 hours. The resulting models are evaluated in our multilingual setting. Since the evaluated trace types are English-only, models trained on them tend to produce reasoning in English at inference time. To mitigate this, we explicitly prompted them to generate responses in the target language of each question, as shown in~\ref{app:prompts}.
Downstream accuracy demonstrates that our traces are superior across all datasets and languages compared to other models (Figure \ref{fig:comp}, bottom-left; detailed results in \ref{app:comp}), with the sole exception of Huatuo traces applied to Qwen on MedMCQA and MedQA.

\noindent \textbf{State-of-the-art performance.}
Finally, we compare the overall performance of the strongest models from prior work - which included both SFT and Reinforcement Learning - 
to our best fine-tuned model based on \texttt{Qwen3-8B}.
The results aggregated by language and dataset are presented in Figure~\ref{fig:comp} (bottom-right), while detailed results can be found in the Appendix, Table~\ref{tab:sota_all}. 
We find that our model in its basic configuration outperforms the best prior model, \textit{Huatuo}, by an average of $+2.9$ points.
In addition, improving our fine-tuned model with two of our traces as examples at inference time obtains the highest overall accuracy, surpassing \textit{Huatuo} by $+7$ points. When considering only the English portion of the datasets, our improved model still outperforms \textit{Huatuo} by $+1.7$ points.

\begin{table*}[]
\centering
  \setlength{\tabcolsep}{4.5pt}
  \renewcommand{\arraystretch}{1.2}
  \small
\begin{tabular}{l|ccc|ccc|ccc|cc}
\toprule
  \textbf{Configuration}   &  \multicolumn{3}{c|}{\textbf{MedExpQA}} & \multicolumn{3}{c|}{\textbf{MedMCQA}} & \multicolumn{3}{c|}{\textbf{MedQA}} & \textbf{AVG}  & \textbf{$\delta$} \\ \midrule
& \textbf{IT}       & \textbf{EN}      & \textbf{ES}      & \textbf{IT}       & \textbf{EN}      & \textbf{ES}   & \textbf{IT}       & \textbf{EN}      & \textbf{ES}     &    \\ \midrule
Llama8B+SFT & 51.6 & 57.4 & 47.3 & 45.5 & 55.2 & 47.3 & 50.4 & 59.6 & 50.2 & \textbf{51.6} &  \\
Llama8B+SFT+reason &  57.4     & 58.0    & 50.0    & 45.6    & 55.0    & 47.4    & 52.0    & 67.4   & 55.8   &  \textbf{54.3 } & \textbf{\textcolor{ForestGreen}{+2.7}}\\
Llama8B+SFT+RAG+reason &  61.7     & 67.0    & 46.2    & 46.8    & 58.8    & 51.5    & 51.6    & 69.8   & 56.7    & \textbf{56.7} & \textbf{\textcolor{ForestGreen}{+5.1}}\\ Llama8B+SFT+RAG+reason+mutlilingual
&  58.0     & 65.4    & 54.3    & 46.8    & 59.0    & 50.6    & 54.0    & 71.6   & 59.1    & \textbf{57.6} & \textbf{\textcolor{ForestGreen}{+6.0}}\\
\bottomrule
\end{tabular}
\caption{\textcolor{black}{Impact of different dimensions at training time. \textbf{Llama8B+SFT}: train on question-answer pairs.
\textbf{Llama8B+SFT+reason}: train on question-reasoning-answer triplets, where the reasoning is distilled from Qwen32B without contextual information from Medical-Wikipedia. 
\textbf{Llama8B+SFT+RAG+reason}: train on question-reasoning-answer triplets for each language separately, where the reasoning is obtained by prompting Qwen32B with relevant contextual information retrieved from Medical-Wikipedia.
\textbf{Llama8B+SFT+RAG+reason+multilingual} same as the previous one, but the model is trained with traces from the three languages at once.}}
\label{tab:ablation}
\end{table*}

\subsection{Ablation on SFT}

To better understand the sources of performance gains, we conduct an ablation study comparing three training settings: \textit{i)} standard SFT on question–answer pairs, and \textit{ii)} SFT on question–reasoning–answer triplets. 
To quantify the contribution of external medical knowledge to the reasoning traces, we generated \textit{iii)} an additional set of reasoning traces without incorporating information from the \textit{Medical-Wikipedia} by prompting \texttt{Qwen3-32B} without retrieving context from the knowledge base. This setup allows us to disentangle the effects of externally injected medical knowledge from those of reasoning capabilities distilled from \texttt{Qwen3-32B} alone. 
Due to the training computational cost, we constrain the analysis to one model (\texttt{Llama-3.1-8B-Instruct}).
Finally, our ablation study includes training the model individually on each language versus training it jointly on all languages simultaneously (see Appendix~\ref{app:multiling} for more details).
As reported in Table~\ref{tab:ablation}, training on reasoning traces yields an average improvement of $+2.7$ accuracy points over standard SFT, while grounding traces in Medical-Wikipedia provides an additional $+3.3$ points gain, showcasing how structured reasoning and medical knowledge from Wikipedia help to improve performance.

\section{Qualitative error analysis}\label{sec:error-analysis}

Although aggregate accuracy metrics summarize model performance, they do not explain the underlying sources of error. To address this limitation, manual error analysis is used to identify systematic failure modes.
First, we analyze a subset of 20 questions that our best-performing model (\texttt{Qwen3-8B}) trained on our traces answered incorrectly. We asked two medical doctors to determine which dimensions would be useful to analyze to help us understand the primary error sources. Based on this sample, they identified that issues were raised because of: \textit{i)} incomplete or incorrect use of information provided in the question, \textit{ii)} insufficient factual medical knowledge, and \textit{iii)} limitations in applying logical reasoning based on that knowledge. Based on these findings, we constructed three questions to guide further analysis:
\emph{i) Does the answer take into account the useful elements present in the question?
ii) Does the answer report medical knowledge mistakes?
iii) Does the answer contain logical mistakes?}

To quantify the contribution of each error source, we randomly sampled an additional 75 question-option-reasoning triplets and reviewed them with the support of eleven medical doctors selected to cover different linguistic areas\footnote{Six from Italy (IT); two from Peru and one from Spain (ES); one from Canada and one from Vietnam (EN).}. The physicians were asked to analyze triplets in their native language (English, Italian, or Spanish). For each triplet, we asked one doctor to analyse it in the light of the three defined questions. 
A careful analysis of the answers revealed that the primary performance bottleneck is the model's limited medical knowledge: the most salient reasoning inconsistencies arise from difficulties in integrating relevant clinical information rather than from flawed reasoning per se. Moreover, overlooked information frequently concerns the interplay between patient characteristics and underlying medical conditions.

More specifically, this includes failure to integrate the full patient history, instead overemphasizing isolated details (Appendix \ref{app:err_an}, Figure~\ref{fig:manual_3}); difficulty in correctly applying standard diagnostic protocols (Appendix \ref{app:err_an}, Figure~\ref{fig:manual_1}); and restricted capacity to exercise nuanced clinical judgment (Appendix \ref{app:err_an}, Figure~\ref{fig:manual_2}). Additional illustrative cases for each language are provided in Appendix \ref{app:err_an}.










\section{Conclusion}



This paper introduces a method for generating multilingual medical reasoning traces based on knowledge extracted from the WikiProject Medicine. Our contributions are threefold: first, we present the first dataset of medical reasoning traces for English, Italian, and Spanish, generated by answering questions from MedQA and MedMCQA; second, comprehensive experiments show that exposure to these traces improves both in-context learning and supervised fine-tuning, consistently achieving state-of-the-art performance; third, we release a multilingual reasoning model alongside three new resources: a collection of 500k+ reasoning traces, translated versions of established medical QA benchmarks, and a manual evaluation by medical doctors assessing the reasoning quality of a randomly sampled subset of the generated traces.

Comparative evaluation against previous work confirms that our approach yields stronger accuracy in multilingual, multiple-choice medical question answering, with gains observed even on English-specific benchmarks. An ablation study quantifies the contributions of reasoning, factual knowledge injection, and multilingual training. A qualitative, expert-driven analysis of our best model further reveals that residual errors stem primarily from limitations in the model's medical knowledge rather than from reasoning failures. Together, these findings underscore the value of knowledge-based multilingual reasoning traces as a scalable path toward more reliable clinical decision-support tools.



\section*{Limitations}

Our work has some limitations. First, we focus on only three languages, namely, English, Italian, and Spanish, whereas multilingual medical reasoning could be extended to a much broader set of languages. Second, we rely on Wikipedia as our primary source of medical knowledge due to its cross-lingual parallelism, though more authoritative medical sources could further strengthen grounding. \textcolor{black}{The use of Wikipedia may introduce Wikipedia-derived biases into the multilingual training, and the possible selection bias introduced by filtering traces solely based on the correctness of final answers.} Furthermore, the English knowledge base is significantly larger than the others due to the nature of Wikipedia pages.
Third, our fine-tuning experiments are restricted to 8B-parameter models because of computational constraints; scaling to larger models remains unexplored. Finally, we limit our study to multiple-choice medical question answering, leaving open-ended clinical reasoning tasks for future work.

\section*{Ethical considerations}
Despite the promising results, our work presents some potential risks. First, reasoning traces may contain factual inaccuracies, and incomplete clinical considerations, even when grounded in external medical knowledge. This is particularly critical in healthcare, where incorrect recommendations could negatively affect clinical decision-making. Our manual analysis represents a first step to assess this dimension.
Second, reasoning traces should not be interpreted as faithful explanations of the internal decision process of the model. While they may improve interpretability, they can also create a false sense of reliability or transparency.

\section*{Acknowledgments}
We acknowledge the contributions of the following medical doctors who contributed to performing the manual error analysis described in Section \ref{sec:error-analysis}: Marcello Albini, Bianca Bernante, Hong Anh Chau, Davide Dotto, Marta Dziech, Daniel García Guadix, Sara Malerba, Elena Maurin, Matteo Savio, José Luis Paredes Sosa, Luca Vavolo, Juan Luis Cuentas Zavala. This work has been partially supported by the Basque Government (Research group funding IT-1805-22), HumanAIze (AIA2025-163322-C61) funded by MICIU/AEI/10.13039/501100011033 and DeepThought (PID2024-159202OB-C21) funded by MICIU/AEI /10.13039/501100011033 and by ERDF, EU.

\bibliography{custom}

\appendix

\section{Related Work on Medical Reasoning} \label{ap:rel_work}

\newtcolorbox{promptbox}{
  colback=gray!8,
  colframe=gray!60,
  boxrule=0.48pt,
  arc=1mm,
  left=2mm, right=2mm, top=1mm, bottom=1mm,
  breakable,
  enhanced,
  halign=flush left,
  fontupper=\ttfamily\small,
}

\citet{chen-etal-2025-towards-medical} proposed \textit{Huatuo}, among the first works to investigate the generation of reasoning traces for medical QA. Huatuo relies on automatic generation of verifiable problems from multiple-choice medical QA datasets. The generated traces, distilled from closed-source LLMs, are used to train models via supervised fine-tuning (SFT) and reinforcement learning (RL) in both Chinese and English. The authors found that models trained on those traces outperform their base counterparts on medical QA, even if the sole source of medical knowledge is a closed-source LLM.

\citet{wu2025medreasonelicitingfactualmedical} advance beyond pure distillation by linking each question with factual medical knowledge. Such knowledge is derived from PrimeKG \citep{Chandak2023}, a knowledge graph of 17k diseases designed for medicine analysis in English. The model trained on those traces (\textit{MedReason}) achieves slightly better performance than Huatuo, highlighting that medically grounded data can enhance reasoning capabilities for the English language.

\citet{huang2025m1unleashpotentialtesttime} further advanced the field focusing on the effect of test-time-scaling \citep{muennighoff2025s1simpletesttimescaling}. The authors train a model on a few thousand examples generated via distillation, highlighting that the quality of the data is what matters.\\
Other recent works follow the approach of generating traces and training on them via SFT or RL.
\citet{sun-etal-2025-reasonmed} build on distillation approaches with architectural adjustments for trace generation, while
\citet{yu2025finemedlmo} construct hundreds of thousands of synthetic medical instructions derived from Common Crawl. While innovative, this method relies heavily on LLM-generated content with limited human validation and uncertain underlying medical quality.
\citet{liu2025distillationpushinglimitsmedical} explore eliciting reasoning capabilities from models without distillation and avoiding intensive use of resources, similarly to \citet{thapa2025disentangling}, who present a method to optimize the use of existing traces. Finally, \citet{wang2025medicalreasoningerallms} provide a comprehensive review of recent advances.

\section{Prompts and Translation}\label{app:prompts_translation}




\begin{table}[ht]
\centering
\small
 \renewcommand{\arraystretch}{1.2}
\begin{tabular}{l|lll|ll}\toprule
         & \multicolumn{3}{c|}{\textbf{Original Split}} & \multicolumn{2}{c}{\textbf{Our Split}} \\  \midrule    
         & \textbf{train}       & \textbf{val}       & \textbf{test}     & \textbf{train}      & \textbf{test}     \\ \midrule
MedExpQA & 434         & 63        & 125      & -                     & 622            \\ 
MedMCQA  & 182k      & 4k      & 6k     & 182k                & 4k           \\ 
MedQA    & 10k       & 1.3k      & 1.3k     & 10k                 & 1.3k   \\ \bottomrule       
\end{tabular}
\caption{Medical QA datasets distribution for both the original split and the one we utilize. We employed MedExpQA only for testing. Following prior work, we used the original training split of both MedQA and MedMCQA for the development of the reasoning traces, and performed the evaluation on the validation split.}
\label{tab:dataset_descr}
\end{table}

\subsection{Prompts} \label{app:prompts}

\noindent \textbf{Prompt template for automatic translation.} The automatic translation is performed in 5-shot. Here we report the prompt and the examples provided for each Language.\\
System prompt:

\begin{promptbox}
\small
You are a helpful assistant that translates text from \{source\_language\} to \{target\_language\}. \\
You must translate the text without adding any additional information or explanations. Never add any additional information or explanations.
\end{promptbox}

\normalsize
User prompt:
\begin{promptbox}
\small
Translate the following text to \{target\_language\}.\\
Return only the translated text.\\{text}\\
\end{promptbox}

\normalsize

Examples:
\small \begin{promptbox}
input: "Growth hormone has its effect on growth through?"\\
output it: "L'ormone della crescita esercita il suo effetto sulla crescita attraverso?"\\
output es: "La hormona del crecimiento tiene su efecto sobre el crecimiento a través de?"\\
input: "What is the main function of the liver?"\\
output it:  "Qual è la funzione principale del fegato?"\\
output es: "Cuál es la función principal del hígado?"\\
input: "Thyroxine"\\
output it: "Tiroxina"\\
output es: "Tiroxina"\\
input: "IG1-1"\\
output it: "IG1-1"\\
output es: "IG1-1"\\
input: "Duodenal Switch"\\
output it: "Duodenal Switch"\\
output es: "Switch Duodenal"\\
\end{promptbox}\normalsize

\noindent \textbf{Prompt template for context rewriting.}

\begin{figure}[t]
  \includegraphics[width=\columnwidth]{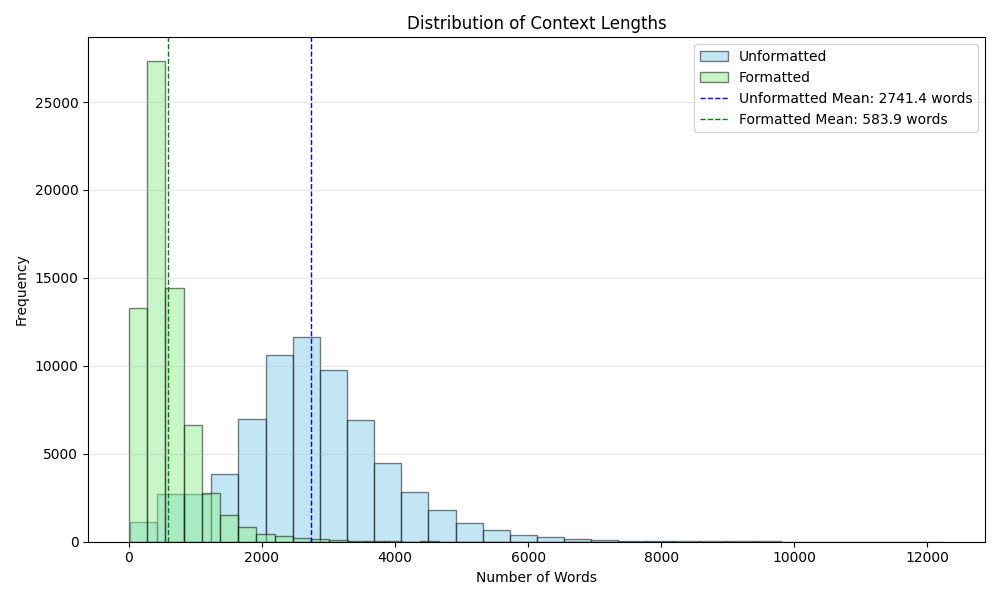}
  \caption{Length distribution of the retrieved context per question versus the LLM-formatted context. The formatted context is notably shorter than the original one.}
  \label{fig:formatted_vs_unformatted_context}
\end{figure}

\small \begin{promptbox}
You are a helpful assistant that rewrites medical text to be more clear and concise.\\
Your task is to rearrange the text to improve its clarity and coherence.\\
Return the rewritten text without saying anything else.
\end{promptbox}\normalsize
The formatting of the context produced the results reported in Figure~\ref{fig:formatted_vs_unformatted_context}.

\noindent \textbf{Prompt template for reasoning trace generation.}

System prompt:
\small \begin{promptbox}
You are a medical expert that reason about medical questions given some provided context. 
\end{promptbox}\normalsize
User prompt:
\small \begin{promptbox}
<question> \{question\}<\/question>\\
<context>\{context\}</context>\\
Please respond to the above question <question> using the Chain of Thought (CoT)  reasoning method, considering the information 
in the <context>.\\
You evaluate and reason on these options and determine a path to the correct answer.<possible answers>\\
\{options\}\\
</possible answers>\\
The correct answer is <answer>\{answer\}</answer>.
\\
1. Explore the question. If any of the possible answer paths seem useful,  incorporate them naturally as if you discovered them yourself.\\
2. Analyze and evaluate the <possible answers> carefully, expanding on the most relevant ones to construct a logical, well-supported explanation.       
3. Do not mention the existence of <context> nor provided <answer> in your response. \\
4. Do not assume the given <answer> is correct. 
Instead, determine the answer solely based on your reasoning.\\
5. If your final conclusion contradicts the given <answer>, acknowledge potential discrepancies (e.g., "Wait, there might be something wrong) and refine your response accordingly.\\
6. The answer should just report the answer's number/identifier\\
\#\#\# Output:\\
Search for potential reasoning traces:
(Find potentially useful reasoning traces. 
It must be a concise list of traces)\\
Reasoning process:
(Step-by-step reasoning process)\\
Conclusion:
(The final answer resulting from your 
reasoning. 
Must be the identifier of the correct option, 
e.g. "1", "2", "3", "4" o "5".)\\
\end{promptbox}\normalsize

\noindent \textbf{Prompt for 2-shot inference (Baseline).}
System prompt:
\small \begin{promptbox}
You are an \{language\} Al assistant that helps people 
find information.
\end{promptbox}\normalsize
User prompt:
\small \small \begin{promptbox}
You are an expert in the medical domain.
Given a medical <Question> and <Possible answers>, 
your task is to select the correct answer. \\
You must return the answer identifier. Do not provide 
any reasoning or explanation.\\
Question: \{question\}\\
Possible Answers: \{options\}\end{promptbox}\normalsize

\noindent \textbf{Prompt for 2-shot inference with reasoning traces.}

System prompt:
\small \begin{promptbox}
You are an \{language\} Al assistant that helps people find information.
\end{promptbox}\normalsize
For the user prompt, here we report the Italian version. We found that providing the output formatting instructions in the target language enforces the generation in such a language.
The Spanish and English prompts are the minimal adaptation of the following:
\small \begin{promptbox}
You are an expert in the medical domain.\\
Given a medical <Question> and <Possible answers>, 
your task is to reason step by step as if you are 
independently determining the possible reasoning 
paths and deriving the correct answer.\\
1. Explore the question and generate a reasoning 
path to answer it.\\
2. You consider and reason on the <Possible answers>
to determine a path to the correct answer.\\
3. Analyze and evaluate the <Possible answers>  
carefully, expanding on the most relevant ones to 
construct a logical, well-supported explanation.\\
4. Generate your own reasoning approach based on 
your expertise.\\
\#\#\# Input:\\
Question: {question}\\
Possible Answers: {options}\\
\#\#\# Output:\\
Ricerca di potenziali tracce di ragionamento:\\
(trovi tracce di ragionamento potenzialmente utili. 
Deve essere una lista concisa di tracce)\\
Processo di ragionamento:\\
(processo di ragionamento fatto step-by-step.)\\
Conclusione:\\
(La risposta finale che deriva dal tuo reasoning. 
Deve essere il numero dell'opzione corretta, ad 
esempio "1", "2", "3", "4" o "5".)\\
Your answer must be in Italian.
\end{promptbox}\normalsize

\noindent \textbf{Prompt for inference with SOTA models.}
\normalsize
Huatuo, MedReason, and m1 need to be prompted to generate the reasoning in the target language. Otherwise, they generate in English no matter what language the question is. Therefore, we designed these minimalist prompts.\\
Italian
\small \begin{promptbox}
Rispondi alla seguente domanda a scelta 
multipla in italiano:\\{question}\\
\{option\_str\}. \\La risposta deve essere 
in italiano!
\end{promptbox}\normalsize
Spanish
\small \begin{promptbox}
Por favor, responda a la siguiente 
pregunta de opción múltiple en 
español:\\\{question\}\\\{option\_str\}. \\¡La respuesta debe 
estar en español!
\end{promptbox}\normalsize

\begin{figure*}[t]
  \includegraphics [width=\textwidth]{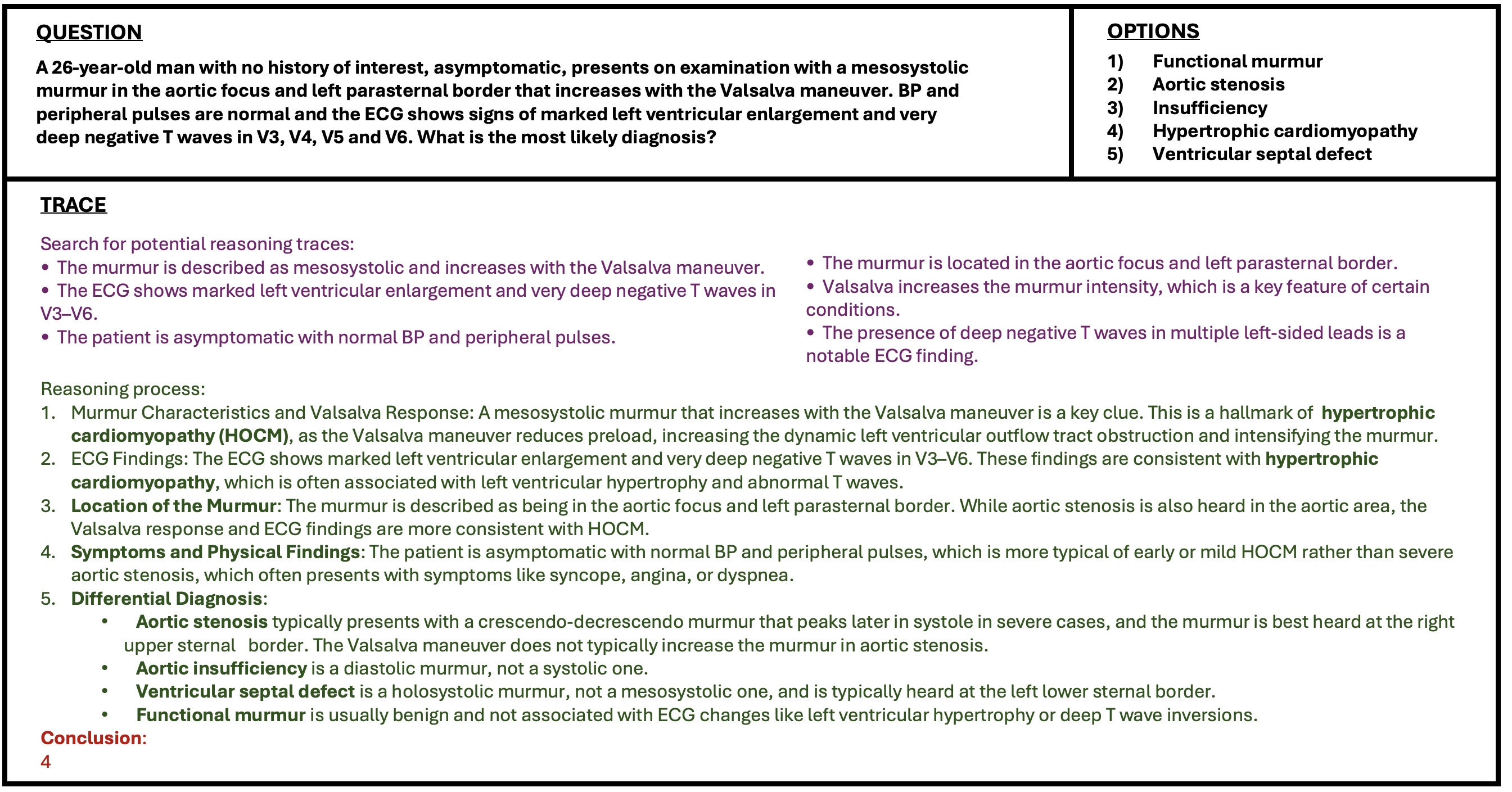}
  \caption{Example of a reasoning trace given a question and the options. The trace is composed of 3 parts. First (\textcolor{violet}{violet}), there is a search for potential paths and useful knowledge required to answer the question, which are then used to elaborate on the provided options (\textcolor{PineGreen}{green} part). Finally (\textcolor{BrickRed}{red}), the final answer is reported as conclusion.}
  \label{fig:example}
\end{figure*}

\subsection{Translation quality} \label{app:back_transl}

In Table~\ref{tab:backtranslation} we report aggregate results. In Tables~\ref{tab:backtranslation_full_it} and \ref{tab:backtranslation_full_es}, we report the scores for back-translation calculated for each dataset split. They are divided into ``options'' and ``questions'', which are the two parts of the datasets we had to translate.

\begin{table}
  \centering
  \small
  \renewcommand{\arraystretch}{1.2}
  \begin{tabular}{l|cccc}
    \toprule
    \textbf{Lang} & \textbf{BERT} & \textbf{COMET} & \textbf{chrF} & \textbf{chrF++} \\
    \midrule
    Italian & 96.5 & 88.7 & 78.0 & 76.3 \\
    Spanish & 96.7 & 89.1 & 79.2 & 77.5 \\
    \bottomrule
  \end{tabular}
  \caption{\label{tab:backtranslation}
    Evaluation metrics for back-translation in Italian and Spanish.
  }
\end{table}

\begin{table*}
  \centering
  \small
  \setlength{\tabcolsep}{3pt} 
  \renewcommand{\arraystretch}{1.3} 
  \begin{tabular}{l|cccc|cccc|l}
    \toprule
     & \multicolumn{4}{c|}{\textbf{Options}} & \multicolumn{4}{c|}{\textbf{Questions}} &  \\
    \midrule
    \textbf{Split} & \textbf{BERT} & \textbf{COMET} & \textbf{chrF} & \textbf{chrF++}  
                   & \textbf{BERT} & \textbf{COMET} & \textbf{chrF} & \textbf{chrF++} & \textbf{AVG}\\
    \midrule
    MedMCQA Test  & 96.9 & 92.6 & 80.3 & 78.5 & 95.5 & 88.3 & 76.9 & 74.5 & \textbf{85.4} \\
    MedMCQA Train & 96.9 & 92.8 & 80.5 & 78.5 & 95.9 & 88.6 & 76.3 & 73.9 & \textbf{85.4} \\
    MedMCQA Val   & 96.8 & 92.0 & 78.6 & 76.8 & 96.0 & 88.0 & 75.7 & 73.4 & \textbf{84.7} \\
    MedQA Test    & 97.7 & 94.1 & 84.9 & 83.5 & 95.7 & 80.4 & 71.1 & 70.0 & \textbf{84.7} \\
    MedQA Train   & 97.7 & 94.1 & 84.7 & 83.3 & 95.7 & 79.9 & 71.0 & 69.8 & \textbf{84.5} \\
    MedQA Val     & 97.7 & 94.1 & 84.6 & 83.1 & 96.0 & 80.9 & 72.7 & 71.5 & \textbf{85.1} \\
    \midrule
    \textbf{Micro AVG} & \textbf{97.0} & \textbf{93.0} & \textbf{82.0} & \textbf{81.0} & \textbf{96.0} & \textbf{84.0} & \textbf{74.0} & \textbf{72.0} & \\
    \bottomrule
  \end{tabular}
  \caption{\label{tab:backtranslation_full_it}
  Back-translation scores on Italian MedMCQA and MedQA (options and questions). Reported scores are  BERTScore, COMET, chrF, and chrF++. The per-column average is weighted by the number of examples per dataset split. }
\end{table*}

\begin{table*}
  \centering
  \small
  \setlength{\tabcolsep}{3pt} 
  \renewcommand{\arraystretch}{1.3} 
  
  \begin{tabular}{l|cccc|cccc|l}
    \toprule
     & \multicolumn{4}{c|}{\textbf{Options}} & \multicolumn{4}{c|}{\textbf{Questions}} & \\
    \midrule
    \textbf{Split} & \textbf{BERT} & \textbf{COMET} & \textbf{chrF} & \textbf{chrF++} 
                   & \textbf{BERT} & \textbf{COMET} & \textbf{chrF} & \textbf{chrF++} & \textbf{AVG} \\
    \midrule
    MedMCQA Test  & 97.3 & 93.7 & 82.3 & 80.5 & 95.6 & 88.9 & 79.0 & 76.5 & \textbf{86.7} \\
    MedMCQA Train & 97.2 & 93.7 & 82.6 & 80.7 & 96.2 & 89.3 & 80.0 & 77.6 & \textbf{87.2} \\
    MedMCQA Val   & 97.2 & 93.1 & 80.8 & 79.1 & 96.2 & 88.6 & 78.1 & 75.8 & \textbf{86.1} \\
    MedQA Test    & 98.0 & 94.9 & 87.0 & 85.7 & 97.4 & 83.7 & 81.2 & 80.3 & \textbf{88.5} \\
    MedQA Train   & 98.0 & 94.7 & 86.6 & 85.3 & 97.5 & 83.9 & 82.1 & 81.1 & \textbf{88.7} \\
    MedQA Val     & 98.1 & 94.8 & 86.7 & 85.3 & 97.6 & 84.4 & 82.8 & 81.9 & \textbf{89.0} \\
    \midrule
    \textbf{Micro AVG} & \textbf{98.0} & \textbf{94.0} & \textbf{84.0} & \textbf{83.0} & \textbf{97.0} & \textbf{86.0} & \textbf{80.0} & \textbf{79.0} & \\
    \bottomrule
  \end{tabular}
  \caption{ \label{tab:backtranslation_full_es} Back-translation scores on Spanish MedMCQA and MedQA (options and questions). Reported scores are  BERTScore, COMET, chrF, and chrF++. The per-column average is weighted by the number of examples per dataset split. }
\end{table*}

\section{Wikipedia Sections excluded by the KB}\label{app:wikipedia_excluded_sections}

Here is the list of Wikipedia sections that have been removed from each page before creating the KB. The scope is to drop unrelevant and noisy sections.
\\
\textbf{it} : [`Note',`Altri progetti', `Voci correlate', `Collegamenti esterni', `Bibliografia', `Nella cultura di massa', `Storia']\\
\textbf{es}:  [`Referencias', `Referencia', `Véase también', `Bibliografía complementaria', `Lectura adicional', `Referencias y notas de pie', `Enlaces externos', `Historia', `Bibliografía', `Notas',  `En la cultura popular', `Otras lecturas', `Ver también', `Bibliografía adicional']\\
\textbf{en}:  [`References', `Additional bibliography', `References and footnotes', `External links', `See also', `History', `Notes', `Bibliography', `Cultural references', `Related articles', `Related topics', `Literature', `Society and culture', `Further reading']

\begin{table*}[!h]
  \centering
  \setlength{\tabcolsep}{4.5pt}
  \renewcommand{\arraystretch}{1.2}
  \small
  \begin{tabular}{l|ccc|ccc|c}
    \toprule
     & \multicolumn{3}{c|}{\textbf{MedExpQA}} 
     & \multicolumn{3}{c|}{\textbf{MedQA}} 
     & \textbf{AVG}\\
    \hline
    \textbf{model} 
    & \textbf{EN} & \textbf{IT} & \textbf{ES}
    & \textbf{EN} & \textbf{IT} & \textbf{ES} &\\
    \midrule

    Llama-3.2-1B-Instruct &
    28.7 \textcolor{BrickRed}{-4.8} &
    20.2 \textcolor{BrickRed}{-8.5} &
    28.6 \textcolor{BrickRed}{-2.2} &
    31.7 \textcolor{BrickRed}{-2.4} &
    21.4 \textcolor{BrickRed}{-8.8} &
    25.3 \textcolor{BrickRed}{-4.6} &
    26.0 \textcolor{BrickRed}{-5.2}\\

    \midrule

    Qwen3-1.7B &
    40.4 \textcolor{ForestGreen}{+6.9} &
    33.9 \textcolor{ForestGreen}{+4.6} &
    25.4 \textcolor{BrickRed}{-0.7} &
    50.4 \textcolor{ForestGreen}{+15.0} &
    35.3 \textcolor{ForestGreen}{+3.1} &
    37.8 \textcolor{ForestGreen}{+4.4} &
    37.2 \textcolor{ForestGreen}{+5.6}\\

    \midrule

    gemma-3-4b-it &
    50.0 \textcolor{ForestGreen}{+0.0} &
    39.6 \textcolor{BrickRed}{-2.4} &
    47.4 \textcolor{ForestGreen}{+9.1} &
    51.3 \textcolor{ForestGreen}{+4.6} &
    45.0 \textcolor{ForestGreen}{+3.9} &
    46.7 \textcolor{ForestGreen}{+5.3} &
    46.7 \textcolor{ForestGreen}{+3.4}\\

    \midrule

    medgemma-4b-it &
    57.4 \textcolor{ForestGreen}{+5.9} &
    48.1 \textcolor{ForestGreen}{+3.4} &
    48.6 \textcolor{ForestGreen}{+2.4} &
    62.3 \textcolor{ForestGreen}{+8.1} &
    48.1 \textcolor{ForestGreen}{+1.2} &
    49.0 \textcolor{ForestGreen}{+0.9} &
    52.3 \textcolor{ForestGreen}{+3.6}\\

    \midrule

    Llama-3.1-8B-Instruct &
    62.8 \textcolor{ForestGreen}{+5.9} &
    44.8 \textcolor{BrickRed}{-0.9} &
    44.9 \textcolor{BrickRed}{-3.0} &
    66.1 \textcolor{ForestGreen}{+5.5} &
    46.8 \textcolor{ForestGreen}{+0.0} &
    49.2 \textcolor{ForestGreen}{+0.8} &
    52.4 \textcolor{ForestGreen}{+1.4}\\

    \midrule

    Qwen3-8B &
    73.0 \textcolor{ForestGreen}{+9.7} &
    59.0 \textcolor{ForestGreen}{+7.4} &
    59.5 \textcolor{ForestGreen}{+1.5} &
    69.1 \textcolor{ForestGreen}{+6.6} &
    58.1 \textcolor{ForestGreen}{+5.6} &
    59.2 \textcolor{ForestGreen}{+4.5} &
    63.0 \textcolor{ForestGreen}{+5.9}\\

    \midrule

    gemma-3-27b-it &
    77.1 \textcolor{ForestGreen}{+4.8} &
    74.3 \textcolor{ForestGreen}{+5.2} &
    74.6 \textcolor{BrickRed}{-1.5} &
    77.7 \textcolor{ForestGreen}{+8.8} &
    70.9 \textcolor{ForestGreen}{+6.7} &
    70.6 \textcolor{ForestGreen}{+6.4} &
    74.2 \textcolor{ForestGreen}{+5.1}\\

    \midrule

    medgemma-27b-text-it &
    81.4 \textcolor{ForestGreen}{+5.9} &
    79.8 \textcolor{ForestGreen}{+6.4} &
    78.9 \textcolor{ForestGreen}{+2.3} &
    84.3 \textcolor{ForestGreen}{+12.5} &
    78.2 \textcolor{ForestGreen}{+11.2} &
    79.1 \textcolor{ForestGreen}{+11.1} &
    80.3 \textcolor{ForestGreen}{+8.2}\\

    \midrule

    Qwen3-32B &
    80.9 \textcolor{ForestGreen}{+4.8} &
    77.0 \textcolor{ForestGreen}{+3.6} &
    77.3 \textcolor{ForestGreen}{+0.7} &
    81.1 \textcolor{ForestGreen}{+6.5} &
    74.6 \textcolor{ForestGreen}{+4.8} &
    76.0 \textcolor{ForestGreen}{+5.6} &
    77.2 \textcolor{ForestGreen}{+4.3}\\
    \bottomrule
    \toprule
    AVG $\delta$
    &
    \textcolor{ForestGreen}{+4.3}
    &
    \textcolor{ForestGreen}{+2.1}
    & 
    \textcolor{ForestGreen}{+1.0}
    &
    \textcolor{ForestGreen}{+7.2}
    &
    \textcolor{ForestGreen}{+3.1}
    & 
    \textcolor{ForestGreen}{+3.8}
    &
    \textcolor{ForestGreen}{+3.6}
    \\

    \midrule
  \end{tabular}
  \caption{ Performances of models exposed to our traces via in-context learning with \textbf{random sampling} (2-shot examples, each being a question-trace-answer triplet). Accuracy is reported with deltas ($\delta$) from baseline (2-shot examples without traces).\label{tab:main_results_fs_no_rag}
  }
\end{table*}

\section{Comparison between different traces types} \label{app:comp}
Table~\ref{tab:fs_all} reports results on the impact of different types of traces on eleven open source models. Models are exposed to the traces via two-shot examples, where each example is composed of a question, the options, and the trace. The two examples are selected via similarity search in an embedding space, comparing the testing question-options pair against all the training question-options pairs.\\
Table~\ref{tab:ft_all} reports results on the impact of different types of traces on \texttt{Qwen3-8B} and \texttt{Llama-3.1-8B-Instruct} at fine-tuning-time.
\begin{table*}
  \centering
  \small
  \begin{tabular}{l|ccc|ccc|ccc}
    \toprule
    \textbf{model} & \multicolumn{3}{c|}{\textbf{MedExpQA}} & \multicolumn{3}{c|}{\textbf{MedMCQA}} & \multicolumn{3}{c}{\textbf{MedQA}} \\
    \midrule
     & IT & EN & ES & IT & EN & ES & IT & EN & ES \\
    \midrule
    Llama-3.2-1B-Instruct & 23.7 & 34.2 & 27.3 & 30.8 & 36.5 & 31.1 & 27.3 & 33.3 & 28.5 \\
    \quad + m23k & \textcolor{Red}{-2.9} & \textcolor{Red}{-1.4} & \textcolor{Red}{-8.1} & \textcolor{Red}{-1.3} & \textcolor{ForestGreen}{+0.4} & \textcolor{Red}{-0.5} & \textcolor{Red}{-0.5} & \textcolor{ForestGreen}{+5.0} & \textcolor{ForestGreen}{+0.2} \\
    \quad + MedReason & \textcolor{Red}{-12.5} & \textcolor{Red}{-11.0} & \textcolor{Red}{-8.1} & \textcolor{Red}{-12.1} & \textcolor{Red}{-7.1} & \textcolor{Red}{-8.7} & \textcolor{Red}{-10.4} & \textcolor{Red}{-1.0} & \textcolor{Red}{-8.8} \\
    \quad + OUR & \textcolor{ForestGreen}{+2.6} & \textcolor{Red}{-1.2} & \textcolor{ForestGreen}{+0.0} & \textcolor{ForestGreen}{+0.6} & \textcolor{ForestGreen}{+3.8} & \textcolor{ForestGreen}{+2.6} & \textcolor{ForestGreen}{+0.1} & \textcolor{ForestGreen}{+2.5} & \textcolor{ForestGreen}{+3.5} \\
    \midrule
    Qwen3-1.7B & 37.0 & 48.3 & 37.0 & 34.5 & 37.4 & 35.4 & 35.5 & 40.5 & 34.9 \\
    \quad + m23k & \textcolor{ForestGreen}{+15.0} & \textcolor{ForestGreen}{+9.3} & \textcolor{ForestGreen}{+15.8} & \textcolor{ForestGreen}{+9.1} & \textcolor{ForestGreen}{+9.9} & \textcolor{ForestGreen}{+9.2} & \textcolor{ForestGreen}{+16.3} & \textcolor{ForestGreen}{+16.7} & \textcolor{ForestGreen}{+18.0} \\
    \quad + MedReason & \textcolor{Red}{-0.2} & \textcolor{ForestGreen}{+4.5} & \textcolor{ForestGreen}{+11.0} & \textcolor{ForestGreen}{+7.5} & \textcolor{ForestGreen}{+11.7} & \textcolor{ForestGreen}{+8.0} & \textcolor{ForestGreen}{+14.9} & \textcolor{ForestGreen}{+19.3} & \textcolor{ForestGreen}{+18.1} \\
    \quad + OUR & \textcolor{ForestGreen}{+8.1} & \textcolor{ForestGreen}{+3.6} & \textcolor{ForestGreen}{+4.7} & \textcolor{ForestGreen}{+16.9} & \textcolor{ForestGreen}{+19.8} & \textcolor{ForestGreen}{+18.0} & \textcolor{ForestGreen}{+11.3} & \textcolor{ForestGreen}{+16.7} & \textcolor{ForestGreen}{+13.4} \\
    \midrule
    gemma-3-4b-it & 49.7 & 50.7 & 55.6 & 42.7 & 46.8 & 42.8 & 43.5 & 49.5 & 44.5 \\
    \quad + m23k & \textcolor{Red}{-4.1} & \textcolor{Red}{-0.3} & \textcolor{Red}{-7.6} & \textcolor{ForestGreen}{+6.6} & \textcolor{ForestGreen}{+5.1} & \textcolor{ForestGreen}{+6.5} & \textcolor{ForestGreen}{+6.5} & \textcolor{ForestGreen}{+9.1} & \textcolor{ForestGreen}{+6.9} \\
    \quad + MedReason & \textcolor{ForestGreen}{+5.5} & \textcolor{ForestGreen}{+6.9} & \textcolor{Red}{-5.2} & \textcolor{ForestGreen}{+4.1} & \textcolor{ForestGreen}{+5.5} & \textcolor{ForestGreen}{+6.0} & \textcolor{ForestGreen}{+10.8} & \textcolor{ForestGreen}{+12.3} & \textcolor{ForestGreen}{+11.8} \\
    \quad + OUR & \textcolor{Red}{-1.9} & \textcolor{ForestGreen}{+11.8} & \textcolor{Red}{-3.1} & \textcolor{ForestGreen}{+11.4} & \textcolor{ForestGreen}{+11.9} & \textcolor{ForestGreen}{+12.3} & \textcolor{ForestGreen}{+7.2} & \textcolor{ForestGreen}{+13.4} & \textcolor{ForestGreen}{+9.1} \\
    \midrule
    medgemma-4b-it & 56.5 & 58.8 & 53.6 & 46.6 & 51.8 & 46.7 & 50.5 & 55.5 & 49.9 \\
    \quad + m23k & \textcolor{Red}{-6.1} & \textcolor{ForestGreen}{+4.4} & \textcolor{ForestGreen}{+5.6} & \textcolor{ForestGreen}{+8.6} & \textcolor{ForestGreen}{+7.9} & \textcolor{ForestGreen}{+8.0} & \textcolor{ForestGreen}{+8.8} & \textcolor{ForestGreen}{+11.2} & \textcolor{ForestGreen}{+10.2} \\
    \quad + MedReason & \textcolor{Red}{-3.7} & \textcolor{Red}{-17.2} & \textcolor{Red}{-8.8} & \textcolor{Red}{-0.9} & \textcolor{Red}{-8.1} & \textcolor{Red}{-2.9} & \textcolor{ForestGreen}{+4.8} & \textcolor{Red}{-3.8} & \textcolor{ForestGreen}{+4.7} \\
    \quad + OUR & \textcolor{Red}{-2.7} & \textcolor{ForestGreen}{+3.1} & \textcolor{ForestGreen}{+5.6} & \textcolor{ForestGreen}{+9.9} & \textcolor{ForestGreen}{+10.7} & \textcolor{ForestGreen}{+11.1} & \textcolor{ForestGreen}{+4.6} & \textcolor{ForestGreen}{+11.8} & \textcolor{ForestGreen}{+7.1} \\
    \midrule
    Llama-3.1-8B-Instruct & 49.2 & 63.0 & 47.9 & 46.4 & 57.5 & 47.9 & 50.4 & 63.1 & 49.7 \\
    \quad + m23k & \textcolor{Red}{-2.8} & \textcolor{ForestGreen}{+0.2} & \textcolor{ForestGreen}{+4.9} & \textcolor{ForestGreen}{+2.7} & \textcolor{ForestGreen}{+2.1} & \textcolor{ForestGreen}{+4.8} & \textcolor{ForestGreen}{+2.4} & \textcolor{ForestGreen}{+5.7} & \textcolor{ForestGreen}{+5.7} \\
    \quad + MedReason & \textcolor{ForestGreen}{+4.4} & \textcolor{Red}{-1.4} & \textcolor{ForestGreen}{+12.9} & \textcolor{ForestGreen}{+1.4} & \textcolor{ForestGreen}{+1.1} & \textcolor{ForestGreen}{+2.3} & \textcolor{ForestGreen}{+7.3} & \textcolor{ForestGreen}{+5.6} & \textcolor{ForestGreen}{+8.6} \\
    \quad + OUR & \textcolor{ForestGreen}{+3.2} & \textcolor{ForestGreen}{+6.4} & \textcolor{ForestGreen}{+2.8} & \textcolor{ForestGreen}{+9.4} & \textcolor{ForestGreen}{+6.0} & \textcolor{ForestGreen}{+10.1} & \textcolor{ForestGreen}{+3.0} & \textcolor{ForestGreen}{+7.8} & \textcolor{ForestGreen}{+7.6} \\
    \midrule
    Qwen3-8B & 60.1 & 66.0 & 64.8 & 53.0 & 58.6 & 53.8 & 57.0 & 67.1 & 60.1 \\
    \quad + m23k & \textcolor{Red}{-0.1} & \textcolor{ForestGreen}{+2.0} & \textcolor{Red}{-3.2} & \textcolor{ForestGreen}{+5.5} & \textcolor{ForestGreen}{+4.4} & \textcolor{ForestGreen}{+4.5} & \textcolor{ForestGreen}{+5.4} & \textcolor{ForestGreen}{+3.8} & \textcolor{ForestGreen}{+4.9} \\
    \quad + MedReason & \textcolor{ForestGreen}{+3.9} & \textcolor{ForestGreen}{+5.2} & \textcolor{ForestGreen}{+0.8} & \textcolor{ForestGreen}{+3.7} & \textcolor{ForestGreen}{+3.9} & \textcolor{ForestGreen}{+4.6} & \textcolor{ForestGreen}{+11.8} & \textcolor{ForestGreen}{+10.3} & \textcolor{ForestGreen}{+9.7} \\
    \quad + OUR & \textcolor{ForestGreen}{+4.6} & \textcolor{ForestGreen}{+8.6} & \textcolor{ForestGreen}{+0.0} & \textcolor{ForestGreen}{+9.3} & \textcolor{ForestGreen}{+8.8} & \textcolor{ForestGreen}{+10.4} & \textcolor{ForestGreen}{+10.5} & \textcolor{ForestGreen}{+8.5} & \textcolor{ForestGreen}{+6.4} \\
    \midrule
    gemma-3-27b-it & 81.9 & 79.7 & 82.2 & 57.9 & 62.8 & 59.4 & 69.2 & 72.9 & 68.9 \\
    \quad + m23k & \textcolor{Red}{-5.9} & \textcolor{Red}{-3.7} & \textcolor{Red}{-7.8} & \textcolor{ForestGreen}{+2.7} & \textcolor{ForestGreen}{+5.0} & \textcolor{ForestGreen}{+5.3} & \textcolor{ForestGreen}{+7.1} & \textcolor{ForestGreen}{+5.4} & \textcolor{ForestGreen}{+6.2} \\
    \quad + MedReason & \textcolor{Red}{-7.5} & \textcolor{Red}{-5.3} & \textcolor{Red}{-6.2} & \textcolor{ForestGreen}{+6.4} & \textcolor{ForestGreen}{+7.5} & \textcolor{ForestGreen}{+6.8} & \textcolor{ForestGreen}{+4.2} & \textcolor{ForestGreen}{+4.5} & \textcolor{ForestGreen}{+3.7} \\
    \quad + OUR & \textcolor{Red}{-7.0} & \textcolor{Red}{-0.3} & \textcolor{Red}{-4.5} & \textcolor{ForestGreen}{+6.0} & \textcolor{ForestGreen}{+6.9} & \textcolor{ForestGreen}{+5.5} & \textcolor{ForestGreen}{+9.4} & \textcolor{ForestGreen}{+8.2} & \textcolor{ForestGreen}{+8.9} \\
    \midrule
    medgemma-27b-text-it & 80.2 & 80.1 & 84.6 & 61.7 & 68.3 & 62.7 & 72.8 & 77.7 & 72.7 \\
    \quad + m23k & \textcolor{Red}{-1.8} & \textcolor{Red}{-7.3} & \textcolor{Red}{-6.2} & \textcolor{ForestGreen}{+5.9} & \textcolor{ForestGreen}{+2.2} & \textcolor{ForestGreen}{+6.3} & \textcolor{ForestGreen}{+5.7} & \textcolor{ForestGreen}{+0.3} & \textcolor{ForestGreen}{+4.9} \\
    \quad + MedReason & \textcolor{Red}{-5.8} & \textcolor{Red}{-0.9} & \textcolor{Red}{-8.6} & \textcolor{ForestGreen}{+6.7} & \textcolor{ForestGreen}{+9.5} & \textcolor{ForestGreen}{+2.7} & \textcolor{ForestGreen}{+1.1} & \textcolor{ForestGreen}{+4.2} & \textcolor{ForestGreen}{+1.2} \\
    \quad + OUR & \textcolor{Red}{-3.4} & \textcolor{ForestGreen}{+4.5} & \textcolor{Red}{-2.7} & \textcolor{ForestGreen}{+9.6} & \textcolor{ForestGreen}{+9.5} & \textcolor{ForestGreen}{+8.8} & \textcolor{ForestGreen}{+8.7} & \textcolor{ForestGreen}{+7.7} & \textcolor{ForestGreen}{+9.8} \\
    \midrule
    Qwen3-32B & 77.4 & 82.7 & 81.1 & 63.9 & 69.0 & 65.1 & 72.3 & 76.5 & 73.9 \\
    \quad + m23k & \textcolor{Red}{-4.6} & \textcolor{ForestGreen}{+2.1} & \textcolor{Red}{-1.9} & \textcolor{ForestGreen}{+3.3} & \textcolor{ForestGreen}{+3.9} & \textcolor{ForestGreen}{+2.8} & \textcolor{ForestGreen}{+3.6} & \textcolor{ForestGreen}{+2.5} & \textcolor{ForestGreen}{+2.9} \\
    \quad + MedReason & \textcolor{ForestGreen}{+1.0} & \textcolor{Red}{-4.3} & \textcolor{ForestGreen}{+2.1} & \textcolor{ForestGreen}{+4.9} & \textcolor{ForestGreen}{+7.0} & \textcolor{ForestGreen}{+6.2} & \textcolor{ForestGreen}{+1.8} & \textcolor{ForestGreen}{+1.9} & \textcolor{ForestGreen}{+0.6} \\
    \quad + OUR & \textcolor{ForestGreen}{+0.2} & \textcolor{Red}{-2.1} & \textcolor{ForestGreen}{+0.8} & \textcolor{ForestGreen}{+4.1} & \textcolor{ForestGreen}{+7.8} & \textcolor{ForestGreen}{+4.5} & \textcolor{ForestGreen}{+5.7} & \textcolor{ForestGreen}{+5.8} & \textcolor{ForestGreen}{+5.7} \\
    \midrule
    Llama-3.3-70B-Instruct & 74.0 & 82.0 & 77.0 & 65.3 & 71.8 & 66.4 & 65.3 & 71.8 & 66.4 \\
    \quad + m23k & \textcolor{ForestGreen}{+3.6} & \textcolor{ForestGreen}{+0.4} & \textcolor{Red}{-4.2} & \textcolor{ForestGreen}{+11.8} & \textcolor{ForestGreen}{+10.6} & \textcolor{ForestGreen}{+10.4} & \textcolor{ForestGreen}{+2.4} & \textcolor{ForestGreen}{+1.1} & \textcolor{ForestGreen}{+2.5} \\
    \quad + MedReason & \textcolor{ForestGreen}{+5.2} & \textcolor{ForestGreen}{+0.4} & \textcolor{ForestGreen}{+7.0} & \textcolor{ForestGreen}{+14.1} & \textcolor{ForestGreen}{+14.8} & \textcolor{ForestGreen}{+14.8} & \textcolor{ForestGreen}{+2.6} & \textcolor{ForestGreen}{+2.3} & \textcolor{ForestGreen}{+3.1} \\
    \quad + OUR & \textcolor{ForestGreen}{+4.9} & \textcolor{ForestGreen}{+1.8} & \textcolor{ForestGreen}{+5.2} & \textcolor{ForestGreen}{+14.7} & \textcolor{ForestGreen}{+13.7} & \textcolor{ForestGreen}{+13.7} & \textcolor{ForestGreen}{+5.7} & \textcolor{ForestGreen}{+4.2} & \textcolor{ForestGreen}{+6.9} \\
    \bottomrule
  \end{tabular}
  \caption{\label{tab:fs_all} Performances of baseline models with deltas produced by exposure to to different types of reasoning traces at inference time. Baselines are models prompted via 2-shot examples, each example being a question-options-answer triplet, while the deltas are obtained by prompting the same models via 2-shot examples, each example being a triplet question-options-trace triplet.}
\end{table*}

\begin{table*}[ht]
  \centering
  \setlength{\tabcolsep}{2pt}
  \renewcommand{\arraystretch}{1.4}
  \small
  \begin{tabular}{l|ccc|ccc|ccc}
    \toprule
    \textbf{Model} & \multicolumn{3}{c|}{\textbf{MedExpQA}} & \multicolumn{3}{c|}{\textbf{MedMCQA}} & \multicolumn{3}{c}{\textbf{MedQA}} \\
    \midrule
     & \textbf{IT} & \textbf{EN} & \textbf{ES} & IT & EN & ES & IT & EN & ES \\
    \midrule
    Llama-3.1-8B-Instruct &
    48.9 & 62.8 & 47.9 & 46.4 & 57.5 & 47.9 & 50.4 & 63.1 & 49.7 \\
    \quad + m23k FT & \textcolor{ForestGreen}{+4.0}
    & \textcolor{ForestGreen}{+4.3} & \textcolor{ForestGreen}{+10.3}&\textcolor{ForestGreen}{+6.8} &\textcolor{ForestGreen}{+3.7} &\textcolor{ForestGreen}{+3.1} &\textcolor{ForestGreen}{+5.9} &\textcolor{ForestGreen}{+10.0} &\textcolor{ForestGreen}{+12.0} \\ 
    \quad + MedReason FT &
     \textcolor{ForestGreen}{+3.6}  &
     \textcolor{Red}{-0.7}  &
     \textcolor{ForestGreen}{+5.5}  &
     \textcolor{Red}{-1.1}  &
     \textcolor{Red}{-3.9}  &
     \textcolor{Red}{-0.6}  &
     \textcolor{ForestGreen}{+0.2}  &
     \textcolor{ForestGreen}{+0.5}  &
     \textcolor{ForestGreen}{+5.2}  \\
    \quad + Huatuo FT &
     \textcolor{Red}{-5.2}  &
     \textcolor{ForestGreen}{+0.1}  &
     \textcolor{ForestGreen}{+0.0}  &
     \textcolor{Red}{-2.3}  &
     \textcolor{Red}{-0.3}  &
     \textcolor{Red}{-0.8}  &
     \textcolor{Red}{-2.0}  &
     \textcolor{ForestGreen}{+3.8}  &
     \textcolor{ForestGreen}{+1.3}  \\
    \quad + OUR FT &
     \textcolor{ForestGreen}{+9.1}  &
     \textcolor{ForestGreen}{+2.6}  &
     \textcolor{ForestGreen}{+6.4}  &
     \textcolor{ForestGreen}{+0.4}  &
     \textcolor{ForestGreen}{+1.5}  &
     \textcolor{ForestGreen}{+2.7}  &
     \textcolor{ForestGreen}{+3.6}  &
     \textcolor{ForestGreen}{+8.5}  &
     \textcolor{ForestGreen}{+9.4}  \\
    \midrule
    Qwen3-8B &
    60.1 & 66.0 & 64.9 & 53.0 & 58.6 & 53.8 & 57.0 & 67.1 & 60.1 \\
    \quad + m23k FT
    &\textcolor{Red}{-1.7} &\textcolor{Red}{-2.1} &\textcolor{Red}{-5.7} &\textcolor{Red}{-3.0} &\textcolor{Red}{-2.0} &\textcolor{Red}{-2.7} &\textcolor{Red}{-3.4} &\textcolor{Red}{-1.9} & \textcolor{Red}{-3.9} \\
    \quad + MedReason FT &
     \textcolor{ForestGreen}{+0.8}  &
     \textcolor{Red}{-1.2}  &
     \textcolor{Red}{-1.9}  &
     \textcolor{ForestGreen}{+0.1}  &
     \textcolor{ForestGreen}{+0.0}  &
     \textcolor{Red}{-0.7}  &
     \textcolor{ForestGreen}{+1.6}  &
     \textcolor{ForestGreen}{+0.5}  &
     \textcolor{Red}{-0.2}  \\
    \quad + Huatuo FT &
     \textcolor{ForestGreen}{+4.2}  &
     \textcolor{ForestGreen}{+3.2}  &
     \textcolor{ForestGreen}{+2.1}  &
     \textcolor{ForestGreen}{+3.4}  &
     \textcolor{ForestGreen}{+0.3}  &
     \textcolor{ForestGreen}{+2.2}  &
     \textcolor{ForestGreen}{+8.2}  &
     \textcolor{ForestGreen}{+6.0}  &
     \textcolor{ForestGreen}{+4.2}  \\
    \quad + OUR FT &
     \textcolor{ForestGreen}{+2.1}  &
     \textcolor{ForestGreen}{+3.7}  &
     \textcolor{ForestGreen}{+4.8}  &
     \textcolor{Red}{-0.7}  &
     \textcolor{ForestGreen}{+2.3}  &
     \textcolor{Red}{-0.6}  &
     \textcolor{ForestGreen}{+4.7}  &
     \textcolor{ForestGreen}{+4.6}  &
     \textcolor{ForestGreen}{+4.9}  \\
    \bottomrule
  \end{tabular}
  \caption{\label{tab:ft_all}
   Performances of baseline models with deltas produced by exposure to to different types of reasoning traces at fine-tuning time. Baselines are models prompted via 2-shot examples, each example being a question-options-answer triplet, while the deltas are obtained by fine-tuning the same models on the dataset of traces.}
\end{table*}

\begin{table*}
  \centering
  \setlength{\tabcolsep}{2pt} 
  \renewcommand{\arraystretch}{1.7} 
  \small
  \begin{tabular}{ll|ccc|ccc|ccc|l}
    \toprule
     & & \multicolumn{3}{c|}{\textbf{MedExpQA}} & \multicolumn{3}{c|}{\textbf{MedMCQA}} & \multicolumn{3}{c|}{\textbf{MedQA}} & \textbf{AVG} \\
    \midrule
    \textbf{family} & \textbf{size}  & \textbf{IT} & \textbf{EN} & \textbf{ES} & \textbf{IT} & \textbf{EN} & \textbf{ES} &\textbf{IT} & \textbf{EN} & \textbf{ES} & \\
    \midrule
    HuatuoGPT & 8B & 
    56.4 & 67.0 & 56.3 & 
    51.1 & \underline{64.1} & \underline{54.4} & 
    57.4 & \underline{75.6} & 58.3 & 60.1 \\
    MedReason & 8B & 
    55.1 & 64.3 & 58.4 & 
    49.5 & 61.1 & 50.3 & 
    54.8 & 70.9 & 62.0 & 58.5 \\
    m1 & 7B & 
    48.6 & 62.2 & 48.1 & 
    47.7 & 59.3 & 48.6 & 
    50.8 & 66.9 & 51.5 & 53.7 \\
    \midrule
    Our SFT & 8B & 
    \underline{62.2} & \textbf{69.7} & \textbf{69.7} & 
    \underline{52.3} & 60.9 & 53.2 & 
    \underline{61.7} & 71.7 & \underline{65.0} & \underline{63.0} \\
    Our SFT+ 2-shot & 8B & 
    \textbf{64.0} & \underline{68.1} & \underline{65.0} & 
    \textbf{61.9} & \textbf{67.6} & \textbf{63.8} & 
    \textbf{68.1} & \textbf{76.7} & \textbf{69.1} & \textbf{67.1} \\
    \bottomrule
  \end{tabular}
  \caption{\label{tab:sota_all}
  Performance (accuracy) of state of the art 8B models on MedExpQA, MedMCQA, and MedQA in Italian, English and Spanish. Our proposed SFT model achieves the best performances in the 2-shot settings.}
\end{table*}

\section{Multilingual vs monolingual SFT}
\label{app:multiling}

\begin{figure}[t]
  \includegraphics [width=\columnwidth]{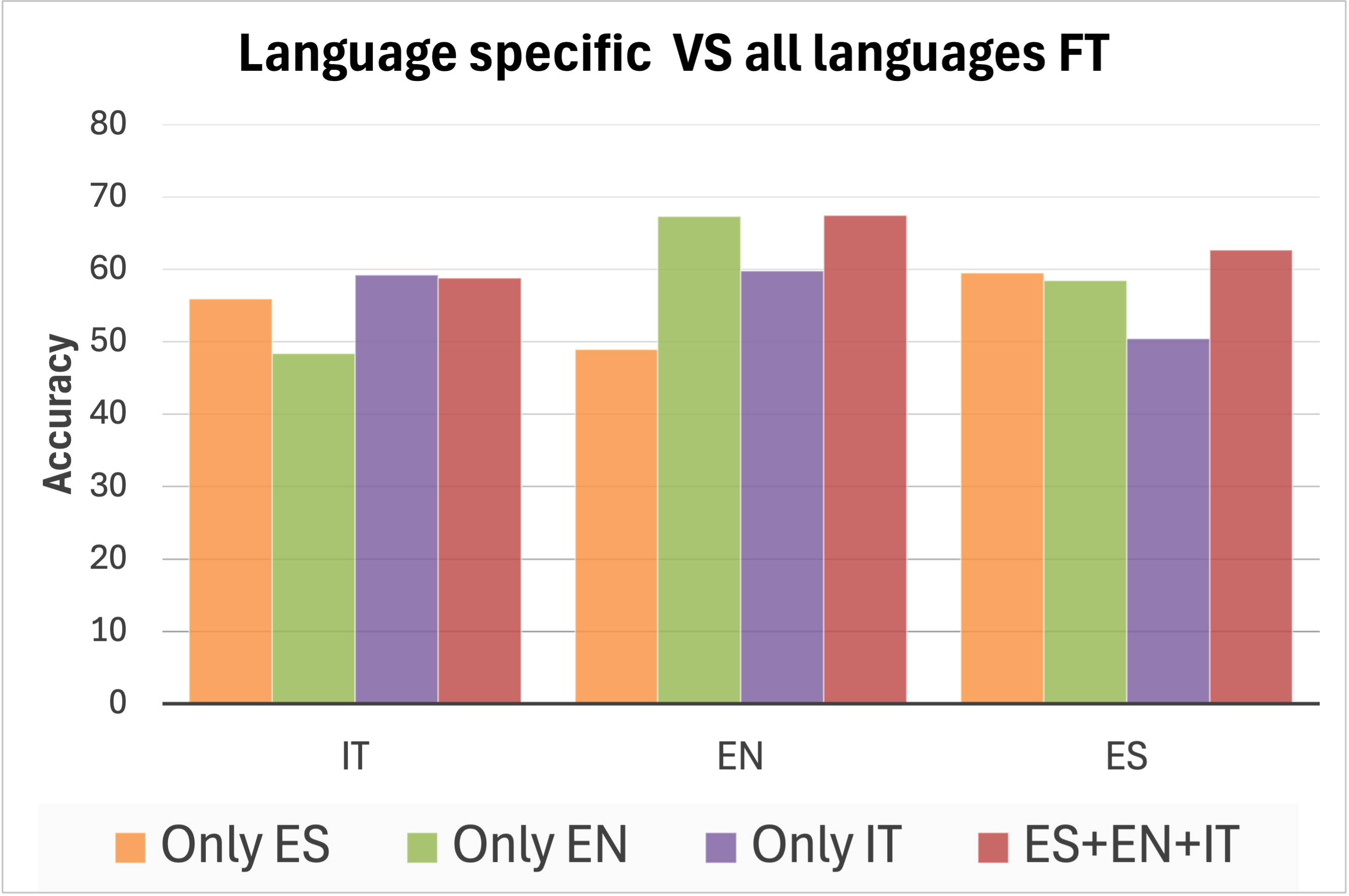}
  \caption{The histogram reports on the impact of
fine-tuning on single-language (‘Only ES/EN/IT”) versus training on all the languages (“ES+EN+IT”)}
  \label{fig:multilingual}
\end{figure}

Experiments demonstrated that integrating multilingual data during fine-tuning leads to improved accuracy on downstream tasks. To further examine this phenomenon, we analyze cross-lingual transfer effects throughout model training. 

We train \texttt{Qwen3-8B} using reasoning traces from individual languages as well as from all three languages put together, following the training protocol outlined in Section~\ref{sec:ft}. The results indicate that multilingual training consistently matches or surpasses the performance of single-language training. In contrast, models trained exclusively on one language exhibit worse accuracy when evaluated on other languages (see Figure~\ref{fig:multilingual}).These findings underscore the positive impact of cross-lingual learning and emphasize the value of incorporating diverse languages into the training data.

\section{Error analysis}\label{app:err_an}
Here we report three examples of the qualitative analysis we performed, one for each language. Each example is provided together with the expert comment on it.

\subsection{English}
\small

QUESTION:

Gustavo comes to the emergency room with skin lesions and general malaise of several days of evolution. He has psoriasiform lesions on the trunk with involvement of palms and soles. He also presents asymmetric non-suppurative joint inflammation and bilateral ocular redness, as well as erosions on the glans penis. In the subsequent anamnesis, Gustavo recognizes a risky sexual contact 20 days before. What is his diagnosis?

OPTIONS:

{`1': `HIV infection.', `2': `Secondary syphilis.', `3': "Reiter`s syndrome.", `4': `Erythema multiforme.', `5': None}

CORRECT ANSWER:

3

MODEL ANSWER:

Search for potential reasoning traces:  
- Psoriasiform lesions on the trunk with involvement of palms and soles.  
- Asymmetric non-suppurative joint inflammation.  
- Bilateral ocular redness.  
- Erosions on the glans penis.  
- History of risky sexual contact 20 days before.  
- These features suggest a sexually transmitted infection with a reactive or disseminated phase.  
- Secondary syphilis is known to present with a rash, mucocutaneous lesions, and systemic symptoms.  
- Reiter's syndrome is a reactive arthritis that follows a genitourinary infection but does not typically present with psoriasiform lesions or mucocutaneous erosions.  
- Erythema multiforme is a hypersensitivity reaction and does not typically involve the glans penis or joints.  
- HIV infection may present with systemic symptoms but does not typically include the described mucocutaneous and ocular features.  

Reasoning process:  
1. Gustavo presents with a combination of skin lesions, joint inflammation, ocular redness, and mucocutaneous erosions.  
2. The psoriasiform rash on the trunk and palms/soles is a key feature of secondary syphilis.  
3. The mucocutaneous erosion on the glans penis is also consistent with secondary syphilis.  
4. The bilateral ocular redness and asymmetric joint inflammation are additional features of secondary syphilis.  
5. The history of a risky sexual contact 20 days before is consistent with the incubation period of syphilis.  
6. Reiter's syndrome is a reactive arthritis that follows a genitourinary infection but does not typically present with psoriasiform lesions or mucocutaneous erosions.  
7. Erythema multiforme is a hypersensitivity reaction and does not typically involve the glans penis or joints.  
8. HIV infection may present with systemic symptoms but does not typically include the described mucocutaneous and ocular features.  

Conclusion:  
2
\normalsize

\noindent \textbf{Expert Comment.}
The model makes a critical medical knowledge error regarding Reiter's syndrome. It incorrectly states that Reiter's does not typically present with psoriasiform lesions or mucocutaneous erosions. In fact, these features are classic characteristics of the syndrome (known as keratoderma blennorrhagicum and circinate balanitis). This fundamental mistake led the model to incorrectly rule out the correct diagnosis.

\subsection{Italian}

\small
QUESTION:
Un uomo di 30 anni è caduto in piedi da un'altezza di 2 metri. All'arrivo al Pronto Soccorso era supino e aveva cateterizzato la vescica con urine chiare. Presenta un gonfiore al tallone sinistro, un leggero formicolio alla parte anteriore della coscia destra e allo scroto e disestesia a entrambi i piedi. Il paziente saluta il barelliere con una stretta di mano e una contemporanea pacca sulla spalla. La rotazione del MMII sulla barella afferrata dalle caviglie è indolore e la compressione del bacino è asintomatica. Il prossimo:

OPZIONI:

{`1': `Il paziente viene fatto sedere sulla barella in modo da poter esplorare il rachide cervicale.', `2': `Fare riferimento alla Rx per lo studio degli arti inferiori.', `3': "Palpare l'addome e ruotarlo in blocco per palpare la colonna vertebrale toracolombare.", `4': `Il paziente verrà sottoposto a un collare cervicale e inviato in radiologia per la diagnostica per immagini del rachide cervicale e degli arti inferiori.', `5': "Verrà richiesta un'ecografia addominale."}

RISPOSTA CORRETTA:

3

RISPOSTA DEL MODELLO:

**Ricerca di potenziali tracce di ragionamento:**  
- Il paziente ha subito una caduta da 2 metri in piedi, con impatto potenzialmente traumatico.  
- All'arrivo al Pronto Soccorso presenta sintomi neurologici: formicolio alla coscia destra, scroto e disestesia ai piedi.  
- La stretta di mano e la pacca sulla spalla suggeriscono una funzionalità motoria conservata, ma non esclude una lesione spinale.  
- Il gonfiore al tallone sinistro potrebbe indicare una frattura o una contusione, ma non spiega i sintomi neurologici.  
- La manovra di rotazione del MMII e la compressione del bacino sono asintomatiche, suggerendo una possibile assenza di lesione spinale a livello lombare o sacrale.  
- I sintomi neurologici (formicolio, disestesia) suggeriscono una possibile lesione spinale a livello toracico o cervicale.  
- La manovra di rotazione del MMII e la compressione del bacino non sono sufficienti per escludere una lesione spinale.  
- È necessario un esame neurologico più approfondito, iniziando con la valutazione del rachide cervicale.  

**Processo di ragionamento:**  
1. Il paziente ha subito una caduta da 2 metri in piedi, un evento che può causare lesioni spinali, specialmente se l'atterraggio non è stato attutito.  
2. I sintomi neurologici (formicolio, disestesia) indicano una possibile lesione spinale, in particolare a livello toracico o cervicale.  
3. La stretta di mano e la pacca sulla spalla suggeriscono una funzionalità motoria conservata, ma non escludono una lesione spinale.  
4. La manovra di rotazione del MMII e la compressione del bacino sono asintomatiche, suggerendo una possibile assenza di lesione spinale a livello lombare o sacrale.  
5. Per escludere una lesione spinale a livello cervicale, è necessario effettuare una valutazione neurologica iniziale, iniziando con la valutazione del rachide cervicale.  
6. Il paziente deve essere posizionato in modo da permettere una valutazione neurologica completa, iniziando con il rachide cervicale.  
7. L'opzione 1 propone di far sedere il paziente per esplorare il rachide cervicale, il che è coerente con la valutazione iniziale richiesta.  
8. Le altre opzioni non sono appropriate per la valutazione iniziale di un paziente con sospetta lesione spinale.  

**Conclusione:**  
1
\normalsize

\noindent \textbf{Expert Comment.}
Presenta alcuni errori di conoscenza medica:
Concentra la valutazione sul rachiede cervicale senza senza valutare l'esame neurologico complessivo

Contiene alcuni errori logici:
Non considera la possibilità che la lesione spianale si aggravi facendo mettere seduto io paziente prima di aver controllato tutti i distretti

\subsection{Spanish}

\small

PREGUNTA:

En un paciente con enfermedad pulmonar obstructiva crónica (EPOC) leve en fase estable se evidencia poliglobulia e insuficiencia respiratoria. En la radiografía de tórax no se aprecian alteraciones reseñables. ¿Cuál de los siguientes procedimientos consideraría realizar en primer lugar para descartar la coexistencia de otras enfermedades que puedan justificar los hallazgos descritos?:

OPCIONES:

{'1': `Poligrafía respiratoria.', `2': `Ecocardiograma.', `3': `Tomografía computarizada torácica.', `4': `Gammagrafía pulmonar.', `5': `Punción de médula ósea.'}

RESPUESTA CORRECTA:

1

RESPUESTA DEL MODELO:

Búsqueda de posibles muestras de razonamiento:  
- La poliglobulia en un paciente con EPOC puede deberse a hipoxia crónica, pero también puede ser secundaria a otras condiciones como mieloproliferativas o tumores.  
- La insuficiencia respiratoria en un paciente con EPOC leve sugiere que podría haber una causa subyacente no evaluada.  
- La radiografía de tórax no muestra alteraciones, lo cual no excluye patología pulmonar o sistémica.  
- La gammagrafía pulmonar puede ayudar a evaluar la perfusión y la ventilación pulmonar, lo cual es útil para descartar neumonía, embolia pulmonar o neoplasias.  
- La tomografía computarizada torácica es más sensible que la radiografía, pero no es el primer paso en un paciente con EPOC estable.  
- La ecocardiografía puede ser útil para evaluar hipertensión pulmonar, pero no es el primer procedimiento para descartar otras causas de poliglobulia.  
- La poligrafía respiratoria es útil para evaluar la función pulmonar, pero no aporta información sobre causas sistémicas de poliglobulia.  
- La punción de médula ósea es invasiva y no es el primer paso para descartar causas no hematológicas de poliglobulia.  

Proceso de razonamiento:  
1. El paciente tiene EPOC leve en fase estable, pero presenta poliglobulia e insuficiencia respiratoria, lo cual sugiere una causa subyacente no evaluada.  
2. La poliglobulia puede deberse a hipoxia crónica, pero también puede ser secundaria a otras condiciones como mieloproliferativas, tumores o infecciones.  
3. La gammagrafía pulmonar permite evaluar la perfusión y la ventilación pulmonar, lo cual es útil para descartar neoplasias, embolia pulmonar o neumonía.  
4. La tomografía computarizada torácica es más sensible, pero no es el primer procedimiento a realizar en un paciente con EPOC estable.  
5. La ecocardiografía puede ser útil para evaluar hipertensión pulmonar, pero no es el primer paso para descartar otras causas de poliglobulia.  
6. La poligrafía respiratoria es útil para evaluar la función pulmonar, pero no aporta información sobre causas sistémicas de poliglobulia.  
7. La punción de médula ósea es invasiva y no es el primer paso para descartar causas no hematológicas de poliglobulia.  
8. Por lo tanto, la gammagrafía pulmonar es el procedimiento más adecuado para descartar otras enfermedades que puedan justificar los hallazgos descritos.  

Conclusión:  
4
\normalsize
\noindent \textbf{Expert Comment.}
La respuesta no tiene en cuenta todos los elementos útiles: Información clínica del paciente, usualmente se repetiría el hemograma para ver si sigue teniendo policitemia.

\begin{figure*}[t]
  \includegraphics [width=\textwidth]{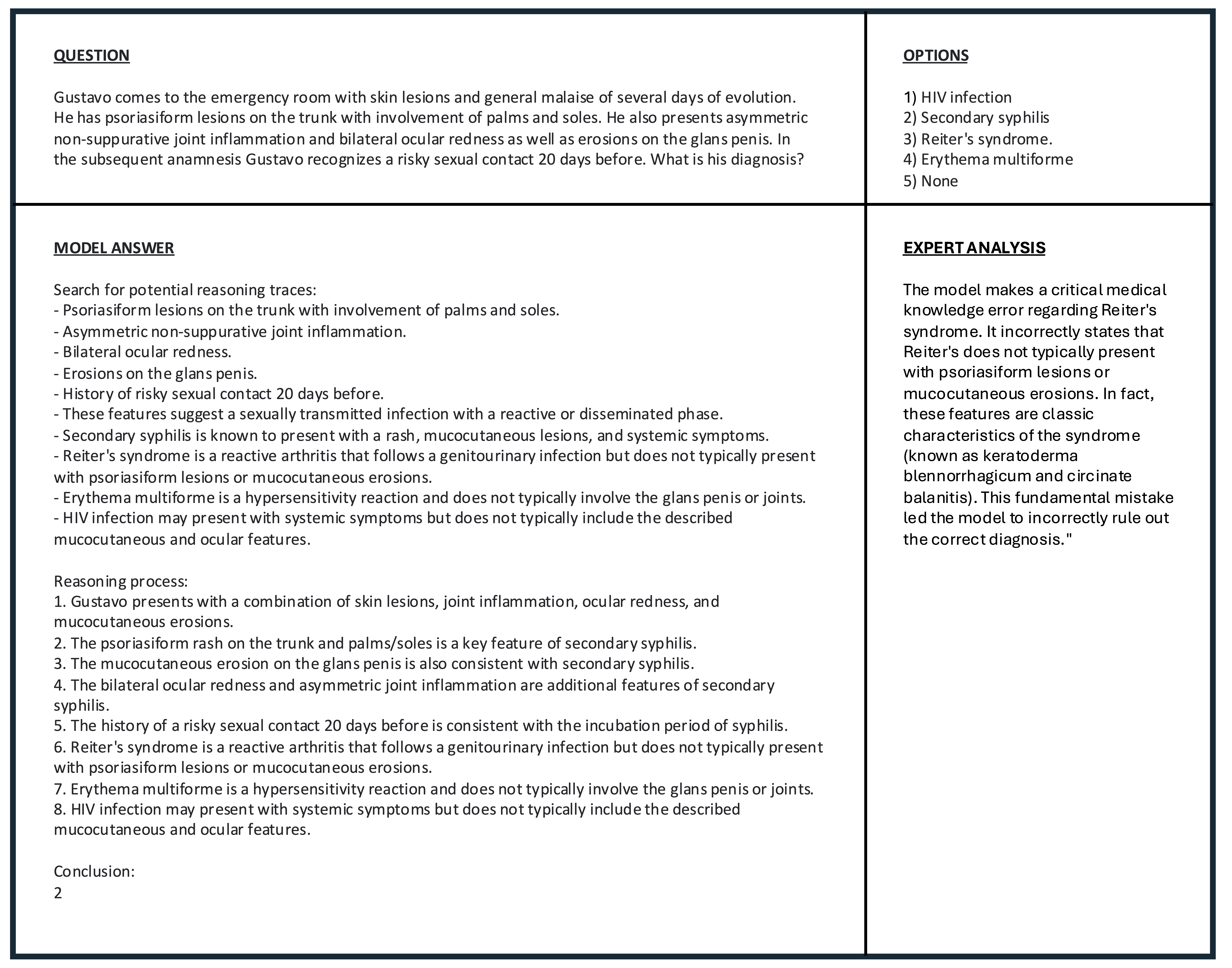}
  \caption{Example of the expert analysis on a model answer which presents a critical clinical knowledge mistake.}
  \label{fig:manual_1}
\end{figure*}

\begin{figure*}[t]
  \includegraphics [width=\textwidth]{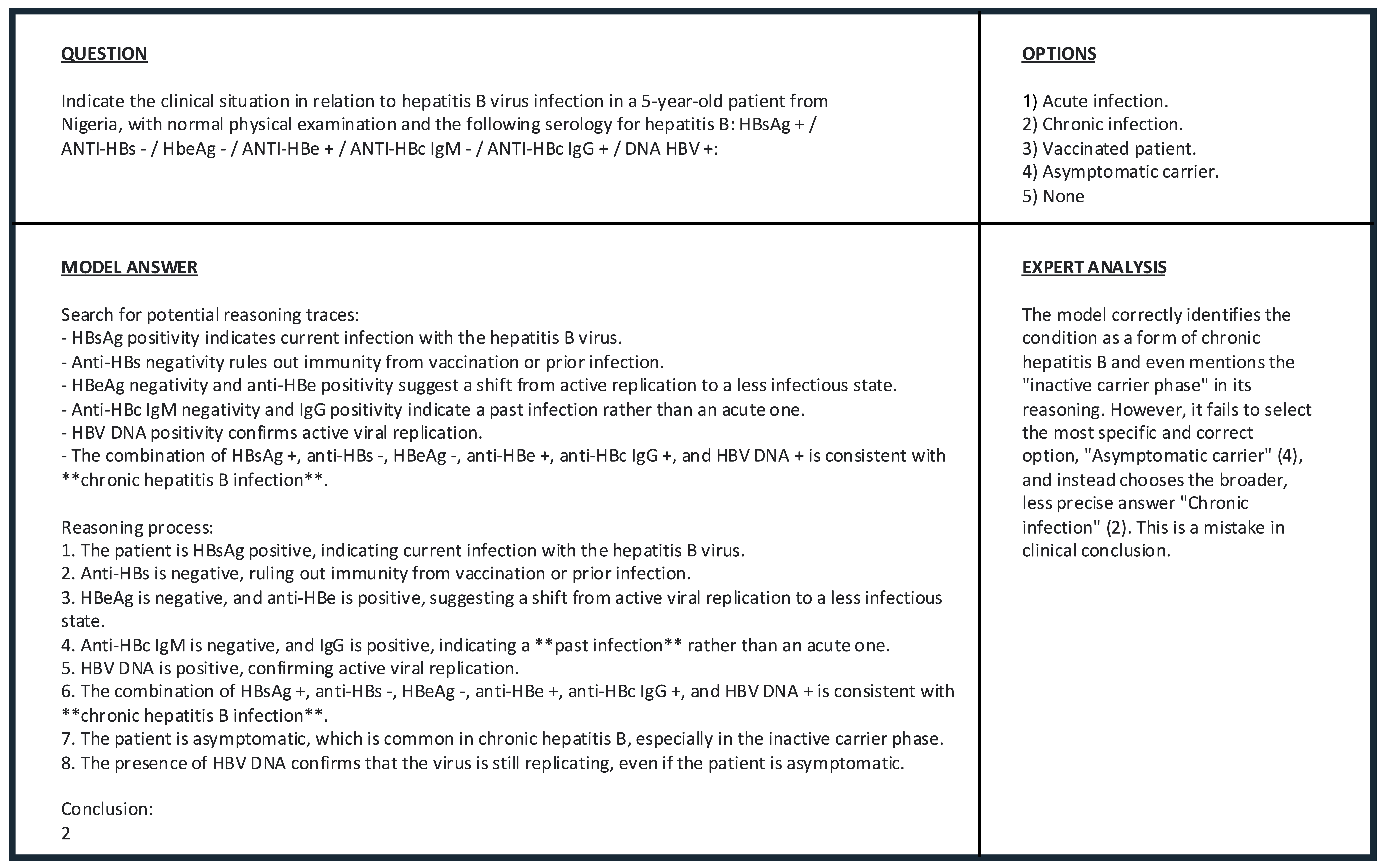}
  \caption{Example of the expert analysis on a model answer which presents a mistake in linking the provided medical information with the most appropriate answer.}
  \label{fig:manual_2}
\end{figure*}

\begin{figure*}[t]
  \includegraphics [width=\textwidth]{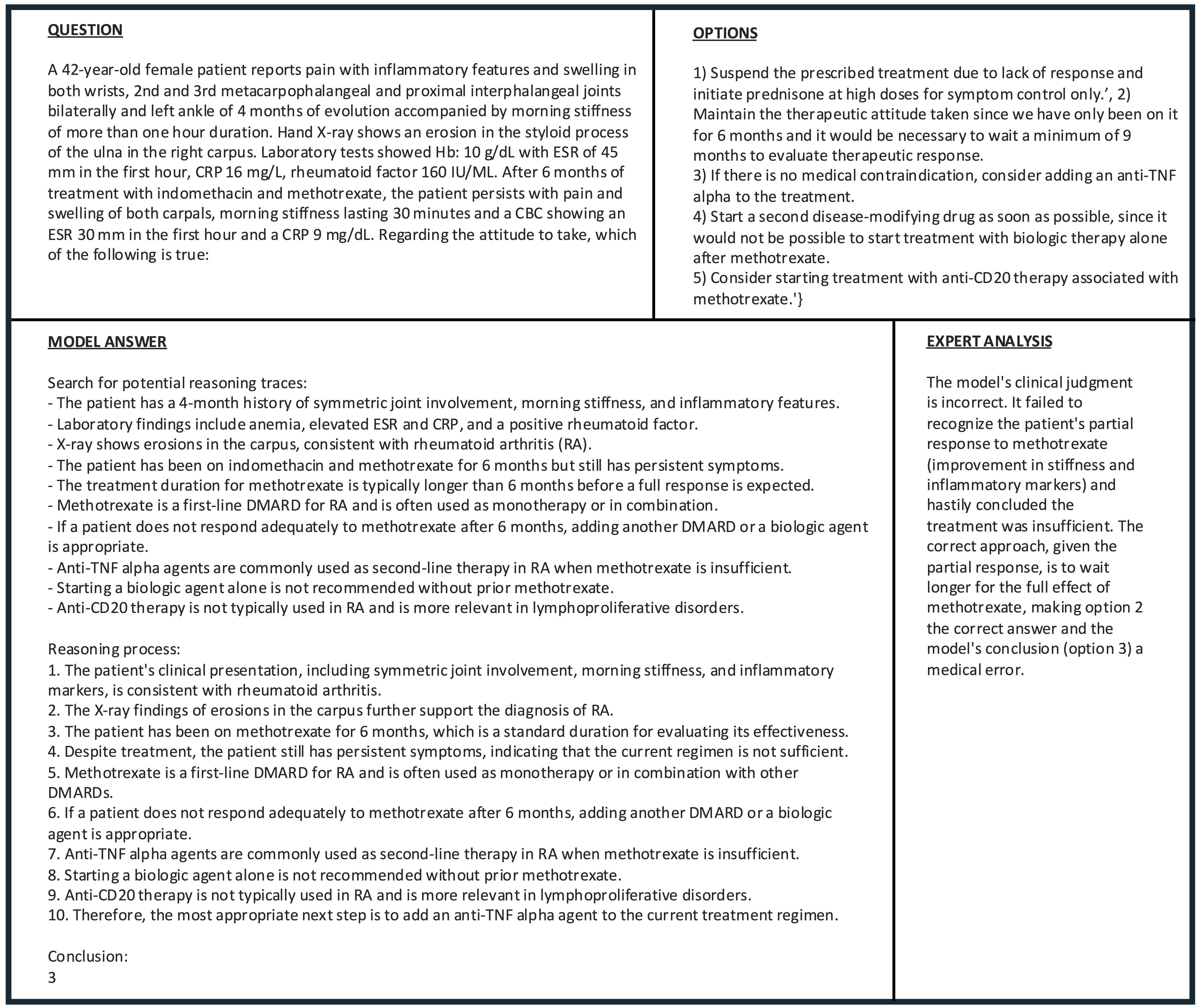}
  \caption{Example of the expert analysis on a model answer which presents misuse of the patient information.}
  \label{fig:manual_3}
\end{figure*}

\end{document}